\documentclass[12pt,journal,onecolumn]{IEEEtran}

\PassOptionsToPackage{numbers, compress}{natbib}

\usepackage{pdfpages}

\usepackage[utf8]{inputenc} 
\usepackage[T1]{fontenc}    
\usepackage{hyperref}       
\usepackage{url}            
\usepackage{booktabs}       
\usepackage{amsfonts}       
\usepackage{nicefrac}       
\usepackage{microtype}      

\usepackage{footnote}
\DeclareFontFamily{U}{dutchcal}{\skewchar\font=45 }
\DeclareFontShape{U}{dutchcal}{m}{n}{<-> s*[1.0] dutchcal-r}{}
\DeclareFontShape{U}{dutchcal}{b}{n}{<-> s*[1.0] dutchcal-b}{}
\DeclareMathAlphabet{\mathlcal}{U}{dutchcal}{m}{n}
\SetMathAlphabet{\mathlcal}{bold}{U}{dutchcal}{b}{n}

\usepackage{graphicx}
\usepackage{booktabs} 
\usepackage{tabularx}

\usepackage{times}
\usepackage{graphicx} 

\usepackage{algorithm}
\usepackage{algorithmic}

\usepackage{hyperref}

\usepackage{times}
\usepackage{epsfig}
\usepackage{graphicx}
\usepackage{amsmath}
\usepackage{amssymb}
\usepackage{booktabs}
\usepackage{etex}
\usepackage{footnote}
\usepackage{tablefootnote}
\usepackage{enumerate}
\usepackage{enumitem}
\usepackage{algorithmic}

\usepackage{graphicx}
\usepackage{latexsym}
\usepackage{amsmath}
\usepackage{amssymb}
\usepackage{subcaption}
\usepackage{multirow}

\usepackage{color}
\usepackage{epstopdf}
\usepackage{array}
\newcolumntype{C}[1]{>{\centering\arraybackslash}m{#1}}

\newcommand{\gr}{{\rm grad}}

\usepackage{url}

\usepackage{amssymb}
\usepackage{amsfonts}
\usepackage{amsthm}
\usepackage{MnSymbol}

\usepackage[cmtip,all]{xy}
\usepackage[all,cmtip]{xy}

\newcommand{\Mi}{\mathcal{M}_{\iota,l}}
\newcommand{\Ci}{\mathcal{C}_{\iota}}
\newcommand{\Oi}{\omega_{\iota}}

\newtheorem{theorem}{Theorem}

\theoremstyle{definition}
\newtheorem{definition}{Definition}
\theoremstyle{definition}

\newtheorem{example}{Example}

\theoremstyle{remark}
\newtheorem{remark}[theorem]{Remark}

\theoremstyle{remark}

\theoremstyle{remark}

\newcommand*{\QEDbs}{\hfill\ensuremath{\blacksquare}}%
\newtheorem{innercustomgeneric}{\customgenericname}
\providecommand{\customgenericname}{}
\newcommand{\newcustomtheorem}[2]{%
	\newenvironment{#1}[1]
	{%
		\renewcommand\customgenericname{\textbf{#2}}%
		\renewcommand\theinnercustomgeneric{\textbf{##1}}%
		\innercustomgeneric
	}
	{\endinnercustomgeneric}
}

\newcustomtheorem{customthm}{Theorem}
\newcustomtheorem{customlemma}{Lemma}
\newcustomtheorem{customcorr}{Corollary}
\newcustomtheorem{customprp}{Proposition}

\usepackage{hyperref}

\begin{document}
	
\title{Fine-grained Optimization of Deep Neural Networks}

\author{Mete Ozay,~\IEEEmembership{Member,~IEEE}}

\IEEEtitleabstractindextext{
\begin{abstract}

In recent studies, several asymptotic upper bounds on generalization errors on deep neural networks (DNNs) are theoretically derived. These bounds are functions of several norms of weights of the DNNs, such as the Frobenius and spectral norms, and they are computed for weights grouped according to either input and output channels of the DNNs. In this work, we conjecture that if we can impose multiple constraints on weights of DNNs to upper bound the norms of the weights, and train the DNNs with these weights, then we can attain empirical generalization errors closer to the derived theoretical bounds, and improve accuracy of the DNNs. 

To this end, we pose two problems. First, we aim to obtain weights whose different norms are all upper bounded by a constant number, e.g. 1.0. To achieve these bounds, we propose a two-stage renormalization procedure; (i) normalization of weights according to different norms used in the bounds, and (ii) reparameterization of the normalized weights to set a constant and finite upper bound of their norms. In the second problem, we consider training DNNs with these renormalized weights. To this end, we first propose a strategy to construct joint spaces (manifolds) of weights according to different constraints in DNNs. Next, we propose a fine-grained SGD algorithm (FG-SGD) for optimization on the weight manifolds to train DNNs with assurance of convergence to minima. Experimental results show that image classification accuracy of baseline DNNs can be boosted using FG-SGD on collections of manifolds identified by multiple constraints.

\end{abstract}}

	\maketitle
	
\IEEEdisplaynontitleabstractindextext
\section{Introduction}
\label{intro}

Despite the practical success of DNNs, understanding their generalization behavior is certainly an open problem \cite{rethinkGen}. The recent theoretical works  \cite{NIPS2017176,NIPS2017204,suzuki18a,zhou18a,arora18b,SizeInd,neyshabur2018a} addressed this problem by extending the early results proposed for shallow linear neural networks (NNs) \cite{Anthony}  for a more general class of DNNs (e.g. NNs with ReLU) and convolutional neural networks (CNNs) (see Table~\ref{tab:compareGen} for a comparison). The proposed asymptotic bounds were obtained by defining matrices of  weights of DNNs using random matrices, and applying concentration inequalities on them. Thereby, the bounds were computed by functions of several $\ell_p$ norms of matrices of weights, where $1 \leq p \leq \infty$. 

In this work, we conjecture that if we can impose multiple constraints on weights of DNNs to set upper bounds of the norms of the weight matrices, and train the DNNs with these weights, then the DNNs  can achieve empirical generalization errors closer to the proposed theoretical bounds, and we can improve their accuracy in various tasks. We pose two problems in order to achieve this goal; (1) renormalization of weights to upper bound norms of their matrices, (2) training DNNs with renormalized weights with assurance to convergence to minima.

\textbf{(1) Bounding norms of weights:} We propose a two-stage renormalization procedure. First, we normalize weights according to the Euclidean, Frobenius and spectral norm, since they are used in the bounds of generalization errors \cite{NIPS2017176,NIPS2017204,suzuki18a,zhou18a,arora18b,SizeInd,neyshabur2018a}. Second, we aim to reparameterize the normalized weights to set a finite and constant upper bound on the weight matrices. For this purpose, we can use a parameter learning approach as utilized in batch normalization (BN) \cite{BN}. However, such an approach substantially increases running time of DNNs during training. In addition, it is not efficient to estimate the parameters using small number of samples in batch training. Therefore, we reparameterize weights according to (a) geometric properties of weight spaces, and (b) statistical properties of features (standard deviation) on which the weights are applied. 
The proposed reparameterization method enables to set  upper bound of each different norm of weight matrices to 1.0. In addition, the proposed renormalization procedure enables to control variance of weights during training of DNNs, thereby assures that DNNs do not have spurious local minima \cite{xie17a}. Employment of standard deviation in reparameterization also  makes optimization landscapes significantly smoother by bounding amount of change of norms of gradients during training. This property has been recently studied to analyze effect of BN on optimization landscape in \cite{Santurkar}. We use this property  to develop a  new optimization method for weight renormalization in this paper, as explained in the next problem. 

\textbf{(2) Training DNNs with renormalized weights:} We consider two subproblems. (i) First, note that, there is not a single procedure used to normalize weights jointly according to all different norms. Thereby, we normalize weights in groups such that similar or different norms can be used to normalize matrices of weights belonging to each different group. We can mathematically prove that the procedure  proposed to solve the previous problem (1) can set an upper bound for all of the aforementioned norms. However, we do not have a mathematical proof to explain whether weights normalized according a single norm can provide the \text{best} generalization bound, and to determine its type. We examine this question in various experiments in detail in the supp. mat. Experimental results show that training DNNs using a set of groups of weights normalized according to all these different norms achieves the best generalization performance in various tasks. Since we cannot mathematically verify this observation, we conjecture that using a diverse set of weights normalized with different constraints improves the generalization error compared to using weights normalized according to single constraint. We consider mathematical characterization of this property as an open problem. 


 Spaces of normalized weights can be identified by different Riemann manifolds \cite{ooAAAI18}\footnote{Please see the supplemental material for more precise mathematical definitions and examples.}; (i) unit norm weights reside on the sphere $Sp(A_lB_l-1)$, (ii) orthonormal weights belong to the Stiefel manifold $St(A_l,B_l)$, and (iii) weights with orthogonal columns reside on the oblique manifold $Ob(A_lB_l)$, at each $l^{th}$ layer of a DNN. We consider training DNNs using a more general setting employing groups of weights which can be normalized according to different normalization constraints. Group wise operations are implemented by concatenating  weight matrices $\omega_{g,l}^i$ belonging to each $g^{th}$ group by ${\omega_{g,l} = (\omega_{g,l}^1, \omega_{g,l}^2, \ldots,\omega_{g,l}^{\mathfrak{g}})}$,${\forall g = 1,2,\ldots,G_l}$. For the corresponding group, a space of concatenated weights is identified by Cartesian product of manifolds of weights $\omega_{g,l}^i,  i=1,2,\ldots,\mathfrak{g}$. In addition, if we renormalize weights using standard deviation of features obtained at each epoch, then geometry of the manifolds of weights also changes. Therefore, we address the second subproblem (ii) which is optimization on dynamically changing product manifolds of renormalized weights. 

DNNs can be trained with multiple constraints using optimization methods proposed for training shallow algorithms \cite{Pls,Lui2012}, and individual manifolds~\cite{ooAAAI18,RBN}. If we employ these  methods on products of weight manifolds (POMs) to train DNNs, then we observe early divergence, vanishing and exploding gradients due to nonlinear geometry of  product of different manifolds. More precisely, the assumption of a bound on the operator norm of Hessian of geodesics in POMs, which is required for assurance of convergence, fails, while performing Stochastic Gradient Descent (SGD) with backpropagation on product of different weight manifolds. Therefore, a non-increasing bound on the probability of failure of the optimization algorithm cannot be computed, and a convergence bound cannot be obtained. In order to solve these problems, we first propose a mathematical framework to make use of the geometric relationship between weight manifolds determined by different constraints (Section~\ref{sec3}). Then, we suggest an approach for training DNNs using multiple constraints on weights to improve their performance under the proposed framework. To this end, we propose a new algorithm that we call fine-grained stochastic gradient descent (FG-SGD) to train DNNs using POMs. We elucidate geometric properties of POMs to assure convergence of FG-SGD to global minima while training nonlinear DNNs with particular assumptions on their architectures, and to local minima while training a more generic class of nonlinear DNNs. Our contributions are summarized as follows:
\begin{enumerate}[wide, labelwidth=!, labelindent=0pt]
\item DNNs trained using weights renormalized by the proposed method (see Proposition~1 in the supp. mat. for derivation) can achieve tighter bounds for theoretical generalization errors compared to using unnormalized weights. These DNNs do not have spurious local minima \cite{xie17a} (see the next section for a detailed discussion). The proposed scaling method generalizes the scaling method proposed in \cite{w_norm} for weight normalization by incorporating geometric properties of weight manifolds. 


\item We explicate the geometry of weight manifolds defined by multiple constraints in DNNs. For this purpose, we explore the relationship between geometric properties of POMs (i.e. sectional curvature),  gradients computed at POMs (Theorem~1), and those of component manifolds of weights in DNNs in Section~\ref{sec3} (please see Lemma~1 in the supp. mat. for more precise results).  

\item We propose an algorithm (FG-SGD) for optimization on different collections of POMs (Section~\ref{sec3}) by generalizing SGD methods employed on  weight manifolds \cite{ooAAAI18,NIPS2017_7107}. Next, we explore the effect of geometric properties of the POMs on the convergence of the FG-SGD using our theoretical results. In the proof of convergence theorems, we observe that gradients of weights should satisfy a particular normalization requirement and we employ this requirement for adaptive computation of step size of the FG-SGD (see \eqref{grad_norm} in Section~\ref{G-SGDdetails}). To this best of our knowledge, this is first result which also establishes the relationship between norms of weights and norms of gradients for training DNNs. We also provide an example for computation of a step size function for optimization on POMs identified by the sphere (Corollary~2 in the supp. mat.). 

\item We propose a strategy to construct sets of identical and non-identical weight spaces according to their employment in groups on input and output channels in DNNs (Section~\ref{sec2}). In the experimental analyses, we apply this strategy to train state-of-the-art networks (e.g. Resnext \cite{resnext}, Mobilenetv2 \cite{Sandler} and DeepRoots~\cite{Ioanno}) which use well-known weight grouping strategies, such as depth-wise or channel-wise grouping, for efficient implementation of DNNs. The results show that the proposed strategy also improves accuracy of these DNNs.

\item We prove that loss functions of DNNs  trained using the proposed FG-SGD converges to minima almost surely (see Theorem~2 and Corollary~1 in the supp. mat.). To the best of our knowledge, our proposed FG-SGD is the first algorithm performing optimization on different collections of products of weight manifolds to train DNNs with convergence properties.


\end{enumerate}

\section{Construction of Sets of POMs in DNNs}
\label{sec2}

Let ${S=\{s_i= (\mathbf{I}_i,y_i) \}_{i=1}^N}$ be a set of training samples, where $y_i $ is a class label of the $i^{th}$ image $\mathbf{I}_i$. We consider an $L$-layer DNN consisting of a set of tensors $\mathcal{W} = \{\mathcal{W}_l \}_{l=1}^L$, where ${\mathcal{W}_l = \{ \mathbf{W}_{d,l} \in \mathbb{R}^{A_l \times B_l \times C_l} \} _{d=1} ^{D_l}}$, and ${\mathbf{W}_{d,l} = [W_{c,d,l} \in \mathbb{R}^{A_l \times B_l}]_{c=1}^{C_l}}$  is a tensor\footnote{We use shorthand notation for matrix concatenation such that $[W_{c,d,l}  ]_{c=1}^{C_l} \triangleq [W_{1,d,l}, W_{2,d,l}, \cdots,W_{C_l,d,l}]$.} of weight matrices $W_{c,d,l}, \forall {l=1,2,\ldots,L}$, for each $c^{th}$ channel ${c=1,2,\ldots,C_l}$ and each $d^{th}$ weight $d=1,2,\ldots,D_l$. In popular DNNs, weights with $A_l=1$ and $B_l=1$ are used at fully connected layers, and those with $A_l >1$ or $B_l>1$ are used at convolutional layers. At each $l^{th}$ layer, a feature representation $f_l(\mathbf{X}_l;\mathcal{W}_l)$ is computed by compositionally employing non-linear functions by
	\begin{equation}
	f_l(\mathbf{X}_l;\mathcal{W}_{l}) = f_l(\cdot;\mathcal{W}_l) \circ f_{l-1}(\cdot;\mathcal{W}_{l-1}) \circ \cdots \circ f_1(\mathbf{X}_1;\mathcal{W}_{1}),
	\end{equation}
	where $\mathbf{X}_{l} = [ X_{c,l}]_{c=1}^{C_l}$, and $\mathbf{X}_1 := \mathbf{I}$ is an image at the first layer ($l=1$). The $c^{th}$ channel of the data matrix $X_{c,l}$ is convolved with the kernel ${W}_{c,d,l}$ to obtain the $d^{th}$ feature map $ X_{c,l+1} : = q(\hat{X}_{d,l})$ by ${\hat{X}_{d,l} = {W}_{c,d,l} \ast X_{c,l}}, \forall c, d, l$, where $q(\cdot)$ is a non-linear function, such as ReLU\footnote{We ignore the bias terms in the notation for simplicity.}.

Previous works \cite{ooAAAI18,NIPS2017_7107} employ SGD using weights each of which reside on a single manifold\footnote{In this work, we consider Riemannian manifolds of normalized weights defined in the previous section. Formal definitions are given in the supp. mat. \label{footnote1}} at each layer of a DNN. We extend this approach considering that each weight can reside on an individual manifold or on collections of products of manifolds, which are defined next.


\begin{definition}[Products of  weight manifolds and their collections]

Suppose that ${\mathcal{G}_l = \{ \mathcal{M}_{\iota,l}: \iota \in \mathcal{I}_{\mathcal{G}_l} \}}$ is a set of weight manifolds$^{\ref{footnote1}}$ $\mathcal{M}_{\iota,l}$ of dimension $n_{\iota,l}$, which is identified by a set of indices $\mathcal{I}_{\mathcal{G}_l}, \forall {l=1,2,\ldots,L}$. More concretely, $\mathcal{I}_{\mathcal{G}_l}$ contains indices each of which represents an identity number ($\iota$) of a weight that resides on a manifold $\mathcal{M}_{\iota,l}$ at the $l^{th}$ layer. In addition, a subset ${\mathcal{I}_{l}^g \subseteq \mathcal{I}_{\mathcal{G}_l}}, {g =1,2,\ldots,G_l}$, is used to determine a subset $\mathcal{G}^g_l \subseteq \mathcal{G}_l$ of weight manifolds  which will be aggregated to construct a product of weight manifolds (POM).
	Each ${\mathcal{M}_{\iota,l} \in  \mathcal{G}^g_l}$ is called a component manifold of a product of weight manifolds which is denoted by $\mathbb{M}_{g,l}$. A weight $\omega_{g,l} \in \mathbb{M}_{g,l}$ is obtained by concatenating weights belonging to $\mathcal{M}_{\iota,l}$, $\forall \iota \in \mathcal{I}^g_{l}$, using  ${\omega_{g,l} = (\omega_1, \omega_2, \cdots, \omega_{|\mathcal{I}^g_{l}|})}$, where $|\mathcal{I}^g_{l}|$ is the cardinality of $\mathcal{I}^g_{l}$. A $\mathcal{G}_l$ is called a  \textit{collection of POMs}. 
\QEDbs
\end{definition}


We propose three schemes called POMs for input channels (PI), for output channels (PO) and input/output channels (PIO) to construct index sets. Indices of the sets are selected randomly using a hypergeometric distribution without replacement at the initialization of a training step, and fixed in the rest of the training. Implementation details and experimental analyses are given in the supp. mat.


\section{Optimization using Fine-Grained SGD in DNNs}
\label{sec3}

\subsection{Optimization on POMs in DNNs: Challenges and Solutions}

Employment of a vanilla SGD on POMs with assurance to convergence to local or global minima for training DNNs using back-propagation (BP) with collections of POMs is challenging. More precisely, we observe early divergence of SGD, and exploding and vanishing gradients in the experiments, due to the following theoretical properties of collections of POMs:

\begin{itemize}[wide, labelwidth=!, labelindent=0pt]
\item Geometric properties of a POM  $\mathbb{M}_{g,l}$ can be different from those of its component manifolds  $\mathbb{M}_{\iota}$, even if the component manifolds are identical. For example, we observe locally varying curvatures when we construct POMs of unit spheres. Weight manifolds with more complicated geometric properties can be obtained using the proposed PIO strategy, especially by constructing collections of POMs of non-identical manifolds. Therefore, assumption on existence of compact weight subsets in POMs may fail due to locally varying metrics within a nonlinear component manifold and among different component manifolds \footnote{Formal definitions and additional details are given in the supp. mat. to improve readability of the main text.}.
\item When we optimize weights using SGD in DNNs, we first obtain gradients computed for each weight $\omega_{g,l} \in \mathbb{M}_{g,l}$  at the $l^{th}$ layer from the $(l+1)^{st}$ layer using BP.  Then, each weight $\omega_{g,l}$ moves on $\mathbb{M}_{g,l}$ according to the gradient. However, curvatures and metrics of $\mathbb{M}_{g,l}$ can locally vary, and they may be different from those of component manifolds of  $\mathbb{M}_{g,l}$ as explained above. This geometric drawback causes two critical problems. First, weights can be moved incorrectly if we move them using only gradients computed for each individual component of the weights, as popularly employed for the Euclidean linear weight spaces. Second, due to incorrect employment of gradients and movement of weights, probability of failure of the SGD cannot be bounded, and convergence cannot be achieved (see proofs of Theorem~2, Corollary~1 and Corollary~2 for details). In practice, this causes unbounded increase or decrease of values of gradients and weights.
\end{itemize}

In order to address these problems for training DNNs, 
we first analyze the relationship between geometric properties of POMs and those of their component manifolds in the next theorem. 

\begin{remark} 
\label{remark} (See Lemma 1 given in the supp. mat. for the complete proof of the following propositions)
Our main theoretical results regarding geometric properties of POMs are summarized as follows:
\begin{enumerate}[wide, labelwidth=!, labelindent=0pt]
\item A metric defined on a product weight manifold $\mathbb{M}_{g,l}$  can be computed by superposition (i.e. linear combination) of Riemannian metrics of its component manifolds.
\item Sectional curvature of a product weight manifold $\mathbb{M}_{g,l}$ is lower bounded by 0. \QEDbs
\end{enumerate}
\end{remark}

We use the first result (1) for \textit{projection} of Euclidean gradients obtained using BP onto product weight manifolds. More precisely, we can compute \textit{norms of gradients} at weights on a product weight manifold by linear superposition of those computed on its component manifolds in {FG-SGD}. Thereby, we can move a weight on a product weight manifold by (i) retraction of components of the weight on component manifolds of the product weight manifold, and (ii) concatenation of projected weight components in FG-SGD. Note also that some sectional curvatures vanish on a product weight manifold $\mathbb{M}_{g,l}$ by the second result (2). For instance, suppose that each component weight manifold $\mathcal{M}_{\iota,l}$ of $\mathbb{M}_{g,l}$ is a unit two-sphere $\mathbb{S}^2$, ${\forall \iota \in \mathcal{I}_{\mathcal{G}_l}}$. Then, $\mathbb{M}_{g,l}$  has unit curvature along two-dimensional subspaces of its tangent spaces, called two-planes. However, $\mathbb{M}_{g,l}$ has  zero curvature along all two-planes spanning exactly two distinct spheres. In addition, weights can always move according to a non-negative bound on sectional curvature of compact product weight manifolds on its tangent spaces. Therefore, we do not need to worry about varying positive and negative curvatures observed at its different component manifolds. The second result also suggests that learning rates need to be computed adaptively by a function of \textit{norms of gradients} and \textit{bounds on sectional curvatures} at each layer of the DNN and at each epoch of FG-SGD for each weight $\omega$ on each product weight manifold $\mathbb{M}_{g,l}$. We employ these results to analyze convergence of FG-SGD and compute its adaptive step size in the following sections.  

\begin{table*}[ht]
	\centering
	\caption{Comparison of generalization bounds. $\mathcal{O}$ denotes big-O and $\mathcal{\tilde{O}}$ is soft-O. $\delta_{l,F}$, $\delta_{l,2}$, and $\delta_{l,2 \to 1}$ denotes upper bounds of the Frobenius norm $\| \omega_l \|_F \leq \delta_{l,F}$, spectral norm $\| \omega_l \|_2 \leq \delta_{l,2}$ and the sum of the Euclidean norms for all rows $\| \omega_l \|_{2 \to 1} \leq \delta_{l,2 \to 1}$ ($\ell_{2 \to 1}$) of weights $\omega_l$ at the $l^{th}$ layer of an $L$ layer DNN using $N$ samples. Suppose that all layers have the same width $\varpi$, weights have the same length $\mathcal{K}$ and the same stride $\mathfrak{s}$. Then, generalization bounds are obtained for DNNs using these fixed parameters by $\| \omega_l \|_2 = \frac{\mathcal{K}}{\mathfrak{s}}$, $\| \omega_l \|_F = \sqrt{\varpi}$ and $\| \omega_l \|_{2 \to 1} = \varpi$. We compute a concatenated weight matrix $\omega_{g,l} = (\omega_{g,l}^1, \omega_{g,l}^2, \ldots, \omega_{g,l}^{|\mathfrak{g}|})$ for the $g^{th}$ weight group of size $|\mathfrak{g}|, g=1,2,\ldots,G_l, \forall l$  using a weight grouping strategy.  Then, we have upper bounds of norms by $ \| \omega_{g,l}\|_F \leq \delta_{g,l,F} \leq 1$, $\| \omega_{g,l}\|_2 \leq\delta_{g,l,2}  \leq 1$ and $\| \omega_{g,l}\|_{2 \to 1} \leq \delta_{g,l,2 \to 1}  \leq 1, g=1,2,\ldots,G_l$, which are defined in Table~\ref{tab:norms}. }	
	\begin{tabular}{|C{3.25cm}|C{5.6cm}|}
		\toprule
		\toprule
		& \multicolumn{1}{|c|}{\textbf{DNNs} (dynamic group scaling)}\\
		\bottomrule
		Neyshabur et al.~\cite{Neyshabur15}     & $\mathcal{O}\Big( \frac{2^L \prod\limits_{l=1}^{L} \prod\limits_{g=1}^{G_l} \delta_{g,l,F}}{\sqrt{N}} \Big)$ \\
		Bartlett et~al.~\cite{NIPS2017204}   & $\mathcal{\tilde{O}} \Bigg( \frac{\prod\limits _{l=1} ^L \prod\limits _{g=1} ^{G_l} \delta_{g,l,2}}{\sqrt{N}} \Big( \sum \limits_{l=1} ^L \prod \limits _{g=1}^{G_l} (\frac{\delta_{g,l,2 \to 1}}{\delta_{g,l,2}})^{\frac{2}{3}} \Big) ^{\frac{3}{2}} \Bigg)$\\
		Neyshabur et al.~\cite{neyshabur2018a} & $\mathcal{\tilde{O}} \Bigg ( \frac{\prod \limits_{l=1}^{L} \prod \limits_{g=1}^{G_l}\delta_{g,l,2}} {\sqrt{N}} \sqrt{L^2 \varpi \sum\limits_{l=1} ^L \prod \limits_{g=1}^{G_l}\frac{\delta^2_{g,l,F}}{\delta^2_{g,l,2} } } \Bigg)$ \\
		\bottomrule
		\bottomrule
		
	\end{tabular}%
	\label{tab:compareGen}%
\end{table*}%

\begin{table}[ht]
	\centering
	\caption{Comparison of norms of weights belonging to different weight manifolds. Suppose that weights $\omega_{g,l}^i \in \mathbb{R}^{A_{l} \times B_{l}}$ belonging to the $g^{th}$ group of size $|\mathfrak{g}|, g=1,2,\ldots,G_l, \forall l$ have the same size $A_{l} \times B_{l}$ for simplicity, and $\sigma(\omega_{g,l}^i)$ denotes the top singular value of $\omega_{g,l}^i$.  Let $\|\omega^i_{g,l}\|_F$, $\|\omega^i_{g,l}\|_2$, and $\|\omega^i_{g,l}\|_{2 \to 1}$, denote respectively the Frobenius, spectral and $\ell_{2 \to 1}$ norms of the weight $\omega_{g,l}^i$. Then, we have $\|\omega_{g,l}\|_F \geq (\prod \limits_{i=1} ^{|\mathfrak{g}|}\|\omega^i_{g,l}\|_F)^{1/|\mathfrak{g}|}$, $\|\omega_{g,l}\|_2 \geq (\prod \limits_{i=1} ^{|\mathfrak{g}|}\|\omega^i_{g,l}\|_2)^{1/|\mathfrak{g}|}$ and $\|\omega_{g,l}\|_{2 \to 1} \geq (\prod \limits_{i=1} ^{|\mathfrak{g}|} \|\omega^i_{g,l}\|_{2 \to 1})^{1/|\mathfrak{g}|}$.}	
	\begin{tabular}{|C{1.5cm}|C{1.50cm}|C{1.5cm}|C{1.5cm}|}
		\toprule
		\toprule
		Norms & \multicolumn{1}{|c}{(i) {Sphere} } & \multicolumn{1}{|c|}{{(ii) Stiefel}} & \multicolumn{1}{|c|}{{(iii) Oblique}}\\
		\bottomrule
		$\|\omega^i_{g,l}\|_{2}$& $ \sigma(\omega_{g,l}^i)$ & $1.0 $ & $\sigma(\omega_{g,l}^i)$\\
		$\|\omega^i_{g,l}\|_{F}$ & $1.0$ & $(B_l)^{1/2}$ & $(B_l)^{1/2}$ \\
		$\|\omega^i_{g,l}\|_{2 \to 1}$ & $1.0$  & $(B_l)^{1/4}$ & $(B_l)^{1/4}$ \\
		\bottomrule
		\bottomrule								
	\end{tabular}%
	\label{tab:norms}%
\end{table}%

\section{Bounding Generalization Errors using Fine-grained Weights}



Mathematically, norms of concatenated weights $\omega_{g,l}, \forall g$, are lower bounded by products of norms of component weights $\omega_{g,l}^i, \forall i$. We compute norms of weights belonging to each different manifold in  Table~\ref{tab:norms}. Weights are rescaled dynamically at each $t^{th}$ epoch of an optimization method proposed to train DNNs using $\Re_{i,l}^t= \frac{\gamma_{i,l}}{\lambda_{i,l}^t}$, where $\gamma_{i,l} >0$ is a geometric scaling parameter and $\lambda_{i,l}^t$ is the standard deviation of features input to the $i^{th}$ weight in the $g^{th}$ group $\omega_{g,l}^i, \forall i,g$. The scaling parameter $\Re_{i,l}^t$ enables us to upper bound the norms of weights by $1$ (see Table~\ref{tab:compareGen}). Computation of upper bounds are given in Proposition~1 in the supplemental material. The proof strategy is summarized as follows:
\begin{itemize}[wide, labelwidth=!, labelindent=0pt]
\item Let $\mathfrak{b}_{i,l}$ be multiplication of the number of input channels and the size of the receptive field of the unit that employs $\omega_{g,l}^i$, and $\hat{\mathfrak{b}}_{i,l}$ be multiplication of the dimension of output feature maps and the number of output channels used at the $l^{th}$ layer, respectively. Then, geometric scaling $\gamma_{i,l}$ of the weight space of $\omega_{g,l}^i$ is computed by
\begin{equation}
\gamma_{i,l} = \sqrt{\frac{1}{\mathfrak{b}_{i,l}+\hat{\mathfrak{b}}_{i,l}} }.
\label{eq:gamma}
\end{equation}		

\item We can consider that standard deviation of features satisfy $\lambda_{i,l}^t \geq 1$ using two approaches. First, by employing the central limit theory for weighted summation of random variables of features, we can prove that $\lambda_{i,l}^t$ converges to $1$ asymptotically, as popularly employed in the previous works. Second, we can assume that we apply batch normalization (BN) by setting the re-scaling parameter of the BN to $1$. Thereby, we can obtain $\frac{1}{\lambda_{i,l}^t} \leq 1$. By definition, $\gamma_{i,l}^2 < B_l, \forall i,l$. In order to show that $ \sigma(\omega_{g,l}^i)  \leq (\gamma_{i,l})^{-1}, \forall i,l$, we apply the Bai-Yin law \cite{bai1993,BAI1988166}. Thereby, we conclude that norms of concatenated weights belonging to groups given in Table~\ref{tab:compareGen} are upper bounded by $1$, if the corresponding component weights given in Table~\ref{tab:norms} are rescaled by $\Re_{i,l}^t, \forall i,l,t$ during training.
\end{itemize}

Note that scaling by $\Re_{i,l}^t$ computed using \eqref{eq:gamma} is different from the scaling method suggested in  \cite{ooAAAI18} such that our proposed method assures tighter upper bound for norms of weights. Our method also generalizes the scaling method given in \cite{xavier} in two ways. First, we use size of input receptive fields and output feature spaces which determine dimension of weight manifolds, as well as number of input and output dimensions which determine number of manifolds used in groups. Second, we perform scaling not just at initialization but also at each $t^{th}$ epoch of the optimization method. Therefore, diversity of weights is controlled and we can obtain weights uniformly distributed on the corresponding manifolds whose geometric properties change dynamically at each epoch. Applying this property with the results given in \cite{xie17a}, we can prove that NNs applying the proposed scaling have no spurious local minima\footnote{We omit the formal theorem and the proof on this result in this work to focus on our main goal and novelty for optimization with multiple weight manifolds.}. In addition, our  method generalizes the scaling method proposed in \cite{w_norm} for weight normalization by incorporating geometric properties of weight manifolds.

\begin{algorithm}[tb]
	\caption{Optimization using FG-SGD on products manifolds of fine-grained weights.}
	\begin{algorithmic}[1]
		\STATE {\bfseries Input:} $T$ (number of iterations), $S$ (training set), \\ $\Theta$ (set of hyperparameters), $\mathcal{L}$ (a loss function), ${\mathcal{I}^l_{g} \subseteq \mathcal{I}_{\mathcal{G}_l}}, \forall g, l$.
		\STATE {\bfseries Initialization:} Construct a collection of products of weight manifolds~$ \mathcal{G}_l$, initialize re-scaling parameters $\mathcal{R}_l^t$ and initialize weights
		${ \omega_{g,l}^t \in \mathbb{M}_{g,l} }$ with ${\mathcal{I}^l_{g} \subseteq \mathcal{I}_{\mathcal{G}_l}}, \forall m,l$.
		\FOR{each iteration $t=1,2,\ldots,T$}
		\FOR{each layer $l=1,2,\ldots,{L}$}
		
		
		
		\STATE ${
		\gr \mathcal{L}(\omega_{g,l}^{t}) := {\rm \Pi}_{\omega_{g,l}^t}  \Big ( \gr_E \; \mathcal{L}(\omega_{g,l}^{t}),\Theta,\mathcal{R}_l^t \Big)},\forall \mathcal{G}_l$.

		
		\STATE $ v_t := h(\gr \mathcal{L}(\omega_{g,l}^{t}), r(t,\Theta)),\forall \mathcal{G}_l$.
		
		\STATE $
		\omega_{g,l}^{t+1} := \phi_{\omega_{g,l}^t}(  v_t,\mathcal{R}_l^t), \forall \omega_{g,l}^t,\forall \mathcal{G}_l$.

		\ENDFOR
		\ENDFOR
		\STATE {\bfseries Output:} A set of estimated weights $\{\omega_{g,l}^T \}_{l=1}^{{L}}, {\forall g}$.
			\label{alg1}
			
	\end{algorithmic}	
\end{algorithm}

\subsection{Optimization on POMs  using FG-SGD in DNNs}
\label{G-SGDdetails}

An algorithmic description of our proposed fine-grained SGD (FG-SGD) is given in Algorithm~\ref{alg1}. At the initialization of the FG-SGD, we identify the component  weight manifolds $\mathcal{M}_{\iota,l}$ of each product weight manifold $\mathbb{M}_{g,l}$ according to the constraints that will be applied on the weights ${\omega_{\iota} \in \mathcal{M}_{\iota,l}}$ for each $g^{th}$ group at each $l^{th}$ layer\footnote{In the experimental analyses, we use the oblique and the Stiefel manifolds as well as the sphere and the Euclidean space to identify component manifolds $\mathcal{M}_{\iota,l}$. Details are given in the supplemental material.}. For $t=1$, each manifold $\mathcal{M}_{\iota,l}$ is scaled by $\Re_{\iota,l}^{t=1}$ using  $\lambda_{\iota,l}^{t=1}=1, \forall \iota,l$. For $t>1$, each $\mathcal{M}_{\iota,l}$ is re-scaled by $\Re_{\iota,l}^{t} \in \mathcal{R}_l^t$ computing empirical standard deviation $\lambda_{\iota}^{t}$ of features input to each weight of $\mathcal{M}_{\iota,l}$, and $\mathcal{R}_l^t$ is the set of all re-scaling parameters computed at the $t^{th}$ epoch  at each $l^{th}$ layer. When we employ a FG-SGD on a product weight manifold $\mathbb{M}_{g,l}$ each weight $\omega_{g,l}^t \in \mathbb{M}_{g,l}$ is moved on $\mathbb{M}_{g,l}$ in the descent direction of gradient of loss at each $t^{th}$ step of the FG-SGD by the following steps: 


 
	
	\textbf{Line 5 (Projection of gradients on tangent spaces):} The gradient  $\gr_E \; \mathcal{L}(\omega_{g,l}^{t})$, obtained using back-propagation from the upper layer, is projected onto the tangent space ${\mathcal{T}_{\omega^t_{g,l}} \mathbb{M}_{g,l}  = \bigtimes \limits _{\iota \in \mathcal{I}_{g}^l} \mathcal{T}_{\omega^t_{\iota,l}} \mathbb{M}_{\iota,l}}$ to compute $\gr \mathcal{L}(\omega_{g,l}^{t})$ at the weight $\omega_{g,l}^{t}$ using the results given in Remark~\ref{remark}, where $\mathcal{T}_{\omega^t_{\iota,l}} \mathbb{M}_{\iota,l}$ is the tangent space at  ${\omega^t_{\iota,l}}$ on the component manifold $\mathbb{M}_{\iota,l}$ of $\mathbb{M}_{g,l} $.

	\textbf{Line 6 (Movement of weights on tangent spaces):} The weight $\omega^t_{g,l}$ is moved on  $\mathcal{T}_{\omega^t_{g,l}} \mathbb{M}_{g,l}$ using 
	\begin{equation}
	h(\gr \mathcal{L}(\omega_{g,l}^{t}), r(t,\Theta)) = -\frac{r(t,\Theta)}{\mathlcal{r}(\omega_{g,l}^t)}\gr \mathcal{L}(\omega_{g,l}^{t}),
	\label{eq:steps}
	\end{equation}
	where $r(t,\Theta)$ is the learning rate that satisfies
\begin{equation}
\sum_{t=0} ^{\infty}r(t,\Theta) = +\infty \; {\rm and} \; \sum_{t=0} ^{\infty} r(t,\Theta)^2 < \infty,
\label{eq:rate}
\end{equation} 
\begin{equation}
{\mathlcal{r}(\omega_{G^m_l}^t) = \max\{ 1,\Gamma_1^t\}^{\frac{1}{2}}}
\label{grad_norm}
\end{equation}
${\Gamma_1^t = (R_{g,l}^{t})^2 \Gamma_2^t}$, ${R_{g,l}^{t} \triangleq  \| \gr \mathcal{L}(\omega_{g,l}^{t})  \|_2}$ is computed using \eqref{eq:grad_norm}, ${\Gamma_2^t = \max \{(2\rho_{g,l}^{t} + R_{g,l}^{t})^2, (1+\mathfrak{c}_{g,l}(\rho_{g,l}^{t} + R_{g,l}^{t}))\} }$, $\mathfrak{c}_{g,l}$ is the sectional curvature of $\mathbb{M}_{g,l}$, ${\rho_{g,l}^{t} \triangleq \rho(\omega_{g,l}^t, \hat{\omega}_{g,l})} $ is the geodesic distance between $\omega_{g,l}^t$ and a local minima $\hat{\omega}_{g,l}$ on $\mathbb{M}_{g,l}$.

The following result is used for computation of the $\ell_2$ norm of gradients.

\begin{customthm}{1}[Computation of gradients on tangent spaces]
	\label{thm_grads}
	The $\ell_2$ norm $\| \gr \mathcal{L}(\omega_{g,l}^{t})  \|_2$ of the gradient $\gr \mathcal{L}(\omega_{g,l}^{t})$ residing on  $\mathcal{T}_{\omega^t_{g,l}} \mathbb{M}_{g,l}$ at the $t^{th}$ epoch and the $l^{th}$ layer can be computed by		
	\begin{equation}
	\| \gr \mathcal{L}(\omega_{g,l}^{t})  \|_2 = \Big (\sum \limits_{\iota \in \mathcal{I}_{g}^l} \gr \mathcal{L}(\omega_{\iota,l}^{t})^2 \Big)^{\frac{1}{2}},
	\label{eq:grad_norm}
	\end{equation} 	
	where $\gr \mathcal{L}(\omega_{\iota,l}^{t})$ is the gradient computed for $\omega_{\iota,l}^{t}$  on the tangent space $\mathcal{T}_{\omega^t_{\iota,l}} \mathbb{M}_{\iota}$, $\forall {\iota \in \mathcal{I}^{l}_g}$.
	\QEDbs
\end{customthm}

	\textbf{Line 7 (Projection of moved weights onto product of manifolds):} The moved weight located at $v_t$ is projected onto $\mathbb{M}_{g,l}$ re-scaled by $\mathcal{R}_l^t$ using $\phi_{\omega_{g,l}^t}(  v_t,\mathcal{R}_l^t)$ to compute $\omega^{t+1}_{g,l}$, where $\phi_{\omega_{g,l}^t}(  v_t,\mathcal{R}_l^t)$ is an exponential map, or a retraction, i.e. an approximation of the exponential map~\cite{absil_retr}.
%
The function  $\mathlcal{r}(\omega_{g,l}^t)$ used for computing step size in \eqref{eq:steps} is employed as a regularizer to control the change of gradient $\gr \mathcal{L}(\omega_{g,l}^{t})$ at each step of FG-SGD. This property is examined in the experimental analyses in the supp. mat. For computation of $\mathlcal{r}(\omega_{g,l}^t)$, we use \eqref{eq:grad_norm} with Theorem~\ref{thm_grads}. In FG-SGD, weights residing on each POM are moved and projected jointly on the POMs, by which we can employ their interaction using the corresponding gradients considering nonlinear geometry of manifolds unlike SGD methods studied in the literature. 
G-SGD can consider interactions between component manifolds as well as those between POMs in groups of weights. Employment of \eqref{eq:steps} and \eqref{eq:rate} at line 7, and retractions at line 8 are essential for assurance of convergence as explained next. 
\subsection{Convergence Properties of FG-SGD}
Convergence properties of the proposed FG-SGD used to train DNNs are summarized as follows:

\textbf{Convergenge to local minima:} The loss function of a non-linear DNN, which employs the proposed FG-SGD, converges to a local minimum, and the corresponding gradient converges to zero almost surely (a.s.). The formal theorem and proof are given in Theorem~2 in the supplemental material.

\textbf{Convergenge to global minima:} Loss functions of particular DNNs such as linear DNNs, one-hidden-layer CNNs, one-hidden-layer Leaky Relu networks, nonlinear DNNs with specific network structures (e.g. pyramidal networks), trained using FG-SGD, converge to a global minimum a.s. under mild assumptions on data (e.g. being distributed from Gaussian distribution, normalized, and realized by DNNs). The formal theorem and proof of this result are given in Corollary~1 in the supp. mat. The proof idea is to use the property that local minima of loss functions of these networks are global minima under these assumptions, by employing the results given in the recent works \cite{Kawaguchi,BrutzkusG17,SDu,OverParam,Yun,Hardt,Ge,criticalGlobal,raghu17a,nguyen17a}.

\textbf{An example for adaptive computation of step size:} Suppose that $\mathbb{M}_{\iota}$ are identified by ${n_{\iota} \geq 2}$ dimensional unit sphere, or the sphere scaled by the proposed scaling method.  If step size is computed using \eqref{eq:steps} with 
\begin{equation}
{\mathlcal{r}(\omega_{G^m_l}^t) = (\max\{ 1, (R_{G^m_l}^{t})^2(2+R_{G^m_l}^{t})^2 \} })^{\frac{1}{2}},
\label{eq:corr1}
\end{equation}
then the loss function converges to local minima for a generic class of nonlinear DNNs, and to global minima for DNNs characterized in Corollary~1. The formal theorem and proof of this result are given in Corollary~2 in the supp. mat. 

We consider analyzing global convergence properties of FG-SGD using different manifolds for larger class of nonlinear DNNs relaxing these assumptions and conditions as a future work.

\begin{table*}[t]
	\centering
	\caption{Mean $\pm$ standard deviation of classification error (\%) are given for results obtained using Resnet-50/101, SENet-Resnet-50/101, and 110-layer Resnets with constant depth (RCD) on Imagenet.}		
\begin{tabular}{|C{5.5cm}|C{3.50cm}|C{5.5cm}|}
		\toprule
		\toprule		
		\textbf{Model} & \textbf{Imagenet(Resnet-50)} &  \textbf{Imagenet(SENet-Resnet-50)} \\		
		Euc. & {\color{red}  24.73 $\pm$ 0.32} & {\color{red} 23.31$\pm$ 0.55}  \\
	St & 23.77  $\pm$ 0.27 & 23.09  $\pm$ 0.41 \\	
		POMs of St & 23.61  $\pm$ 0.22 &  22.97  $\pm$ 0.29 \\		
PIO (Sp+Ob+St) & 23.04  $\pm$ 0.10 &  22.67  $\pm$ 0.15  \\			
PIO (Sp+Ob+St+Euc.) & {\color{blue} 22.89  $\pm$ 0.08 }&  {\color{blue} 22.53  $\pm$ 0.11} \\			
	\bottomrule
	 (Additional results) & \textbf{Imagenet(Resnet-101)} & \textbf{Imagenet(SENet-Resnet-101)} \\
Euc. &  {\color{red} 23.15 $\pm$ 0.09} &   {\color{red} 22.38 $\pm$ 0.30}  \\		
PIO (Sp+Ob+St) & 22.83 $\pm$ 0.06&  21.93 $\pm$  0.12 	\\
PIO (Sp+Ob+St+Euc.) &  {\color{blue} 22.75 $\pm$ 0.02 }&   {\color{blue} 21.76 $\pm$ 0.09}		\\
		\bottomrule		
		\bottomrule
	\end{tabular}%
	\label{tab:summary}%
\end{table*}%

\section{Experimental Analyses}
\label{sec:exp}
 
We examine the proposed FG-SGD method for training DNNs using  different architectures with different configurations on benchmark datasets for image classification tasks. We provide representative results in Table~\ref{tab:summary} in this main text, and the other results in the supp. mat. Implementation details and analysis of computational complexity of the proposed methods are given in the supplemental material. 
We give accuracy of  DNNs for baseline Euclidean (Euc.), the sphere (Sp), the oblique (Ob) and the Stiefel (St) manifold in Table~\ref{tab:summary}.  POMs of St denotes results for weights employed on all input and output channels residing on a POM of St. PIO (\textit{manifolds}) denotes results for collections of POMs of \textit{manifolds} using PIO.

  
Table~\ref{tab:summary} shows results using the state-of-the-art Squeeze-and-Excitation (SE) blocks \cite{senet} implemented for Resnets with 50 layers (Resnet-50) on Imagenet. We run the experiments 3 times and provide the average performance. We first observe that PIO boosts the performance of baseline Euc. ({\color{red} 24.73$\%$}) by $1.84\%$ if sets of weights are employed using Euc, Sp, Ob and St ({\color{blue} 22.89$\%$}). We note that the sets computed for Resnet-50 outperform Resnets with 101 layers ({\color{red} 23.15}) by $0.26\%$. SE blocks aim to aggregate channel-wise descriptive statistics (i.e. mean of convolution outputs) of local descriptors of images to feature maps for each channel. In FG-SGD, we use standard deviation (std) of features extracted from each batch and size of receptive fields of units while defining and updating weight manifolds (see Section 3.3 in supp. mat.). Unlike SE blocks, FG-SGD computes statistical and geometric properties for different sets of input and output channels, and used to update weights by FG-SGD. This property helps FG-SGD to further boost the performance. For instance, we observe that collections of manifolds (23.04$\%$ and {\color{blue} 22.89$\%$} error) outperform SENet-Resnet-50 ({\color{red} 23.31}$\%$ error). Although FG-SGD estimates standard deviation using moving averages as utilized in batch normalization \cite{ICML-2015-IoffeS}, SE blocks estimates the statistics using small networks. Therefore, we conjecture that they provide complementary descriptive statistics (mean and std). The experimental results justify this claim such that sets implemented in SENet-Resnet-50 further boost the performance by providing {\color{blue} 22.53$\%$} error. 

\section{Conclusion and Discussion}
\label{sec:conc}

We introduced and elucidated a problem of training CNNs using multiple constraints  employed on  convolution weights with convergence properties. Following our theoretical results, we proposed the FG-SGD algorithm and adaptive step size estimation methods for optimization on collections of POMs that are identified by the constraints. The experimental results show that our proposed  methods can improve   convergence properties and classification performance of CNNs. Overall, the results show that employment of collections of POMs using FG-SGD can boost the performance of various different CNNs on benchmark datasets. We consider a research direction for investigating how far local minima are from global minima in search spaces of FG-SGD using products of weight manifolds with nonlinear DNNs and their convergence rates. We believe that our proposed framework will be useful and inspiring for researchers to study geometric properties of parameter spaces of deep networks, and to improve our understanding of deep feature representations.

\section*{Supplemental Material}

\begin{appendices}

\section{Bounding Generalization Errors using Fine-grained Weights}

\begin{customprp}{1}[Bounding norms of weight matrices and generalization errors of DNNs]
	\label{prp1}
	Suppose that DNNs given in Table~\ref{tab:compareGen} are trained using weights renormalized by the renormalization method proposed in the main text according to the Frobenius, spectral and column/row wise norms with reparameterization parameters $\Re_{i,l}^t, \forall i,l,t$ with $\lambda_{i,l}^t \geq 1$. Then, norms of renormalized weight matrices are upper bounded by a constant number, and generalization errors of the corresponding DNNs are asymptotically bounded as given in the rightmost column of the Table~\ref{tab:compareGen}, denoted by \textbf{DNNs} (our proposed reparameterization).
	
\end{customprp}
\begin{proof}

	Suppose that matrices of weights $\omega_{g,l}^i \in \mathbb{R}^{A_{l} \times B_{l}}$ belonging to the $g^{th}$ group of size $|\mathfrak{g}|, {g=1,2,\ldots,G_l}$, $\forall l$ have the same size $A_{l} \times B_{l}$ for simplicity, and $\sigma(\omega_{g,l}^i)$ denotes the top singular value of $\omega_{g,l}^i$.  Let $\|\omega^i_{g,l}\|_F$, $\|\omega^i_{g,l}\|_2$, and $\|\omega^i_{g,l}\|_{2 \to 1}$, denote respectively the Frobenius, spectral and $\ell_{2 \to 1}$ norms of the weight $\omega_{g,l}^i$. We note that, matrices of weights $\omega_{g,l}^i$ belonging to the $g^{th}$ group are concatenated by ${\omega_{g,l} = (\omega_{g,l}^1, \omega_{g,l}^2, \ldots,\omega_{g,l}^{\mathfrak{g}})}$, ${\forall g = 1,2,\ldots,G_l}$, to perform group-wise operations in DNNs. 
	Thereby, we can employ bounds for norms of each concatenated matrix  in generalization error bounds given in the leftmost column of Table~\ref{tab:compareGen}, denoted by \textbf{DNNs} (bounds on norms), and obtain the bounds given in the rightmost column of the Table~\ref{tab:compareGen}, denoted by \textbf{DNNs}(our proposed reparameterization). 
	
	We compute norms of matrices of normalized weights $\omega_{g,l}^i$ belonging to each different manifold in  Table~\ref{tab:norms}. These norms are computed using simple matrix calculus considering definitions of matrices residing on each manifold according to the definition given in Table~\ref{tab:manifolds}. From these calculations given in  Table~\ref{tab:norms}, we observe that, the maximum of norm values that a weight $\omega^i_{g,l}$ belonging to the sphere can achieve is $\mathbb{M}_{sp}(\omega^i_{g,l})= \sigma(\omega_{g,l}^i)$, that of a weight belonging to the Stiefel manifold is $\mathbb{M}_{st}(\omega^i_{g,l})=(B_l)^{1/2}$, and that of a weight belonging to the oblique manifold is ${\mathbb{M}_{ob}(\omega^i_{g,l})=\max\{(B_l)^{1/2},\sigma(\omega_{g,l}^i) \} }$.

	In our proposed renormalization method, we first normalize each weight matrix such that the norm of the matrix $\omega^i_{g,l}$ can have one of these values $\mathbb{M}_{sp}(\omega^i_{g,l})$, $\mathbb{M}_{st}(\omega^i_{g,l})$ and $\mathbb{M}_{ob}(\omega^i_{g,l})$. Therefore, we need to reparameterize weight matrices such that norm of each reparameterized weight is less than 1.0. For this purpose, we need show that the rescaling of these norm values by $\Re_{i,l}^t$ is upper bounded by 1.0.

	Weights are rescaled dynamically at each $t^{th}$ epoch of an optimization method proposed to train DNNs using $\Re_{i,l}^t= \frac{\gamma_{i,l}}{\lambda_{i,l}^t}$, where $0 < \gamma_{i,l} < 1.0$ is a geometric scaling parameter and $\lambda_{i,l}^t$ is the standard deviation of features input to the $i^{th}$ weight in the $g^{th}$ group $\omega_{g,l}^i, \forall i,g$. By assumption, $\lambda_{i,l}^t \leq 1.0, \forall i,t,l$. By definition, $ B_l \gamma_{i,l}^2 \leq 1.0, \forall i,l$. In order to show that $ \sigma(\omega_{g,l}^i)  \leq (\gamma_{i,l})^{-1}, \forall i,l$, we apply the Bai-Yin law \cite{bai1993,BAI1988166}. Thereby, we conclude that norms of concatenated weights belonging to groups given in Table~\ref{tab:compareGen} are upper bounded by $1$, if the corresponding component weights given in Table~\ref{tab:norms} are rescaled by $\Re_{i,l}^t, \forall i,l,t$ during training of DNNs.
	
	Since norm of each weight matrix $\omega^i_{g,l}$ is bounded by 1.0, their multiplication for all $g=1,2,\ldots,G_l$ and $\forall l$ is also bounded by 1.0. 
	
\end{proof}

\begin{table}[ht]
	\centering
	\caption{Comparison of norms of weights belonging to different weight manifolds.}	
	\begin{tabular}{|C{1.5cm}|C{1.50cm}|C{1.5cm}|C{1.5cm}|}
		\toprule
		\toprule
		Norms & \multicolumn{1}{|c}{(i) {Sphere} } & \multicolumn{1}{|c|}{{(ii) Stiefel}} & \multicolumn{1}{|c|}{{(iii) Oblique}}\\
		\bottomrule
		\\
		$\|\omega^i_{g,l}\|_{2}$& $ \sigma(\omega_{g,l}^i)$ & $1.0 $ & $\sigma(\omega_{g,l}^i)$\\
		\\
		$\|\omega^i_{g,l}\|_{F}$ & $1.0$ & $(B_l)^{1/2}$ & $(B_l)^{1/2}$ \\
		\\
		$\|\omega^i_{g,l}\|_{2 \to 1}$ & $1.0$  & $(B_l)^{1/4}$ & $(B_l)^{1/4}$ \\
		\bottomrule
		\bottomrule								
	\end{tabular}%
	\label{tab:norms}%
\end{table}%

\begin{table*}[h]
	\caption{Embedded weight manifolds $\mathcal{M}_{\iota}$ used for construction of collection of POMs $\mathbb{M}_{G_l}$, $\forall l$, in the experimental analyses. The Frobenius norm of a convolution weight $\omega$ is denoted by $\| \omega \|_F$. The $b^{th}$ column vector of a weight matrix ${\omega} \in \mathbb{R}^{A_l \times B_l}$ is denoted by $\omega_b$. An $B_l \times B_l$ identity matrix is denoted by  $I_{B_l}$. }	
	\vspace{0.5cm}
	\begin{minipage}{\textwidth} 
		
		\begin{center}
			
			\begin{tabular}{cc }
				\hline
				\textbf{Manifolds} & \textbf{Definitions}  \\
				\hline
				
				\\
				
				The Sphere &	$ \mathcal{S}(A_l,B_l) = \{ {\omega} \in \mathbb{R}^{A_l \times B_l}: \| \omega \|_F = 1  \}$   \\
				
				\\
				
				The Oblique & $  \mathcal{OB}(A_l,B_l) =  \{ {\omega} \in \mathbb{R}^{A_l \times B_l}:  \| \omega_b \|_F = 1, \forall b=1,2,\ldots,B_l  \}$    \\ 				
				
				\\
				
				The Stiefel & $ St(A_l,B_l) = \{ {\omega} \in \mathbb{R}^{A_l \times B_l}:   (\omega^{\rm T} {\omega})= I_{B_l}  \}$   \\

				\hline

			\end{tabular}
		\end{center}
	\end{minipage}%
	\vspace{0.2cm}
	\label{tab:manifolds}%
\end{table*}%
\begin{table*}[ht]
	\centering
	\caption{Comparison of generalization bounds. $\mathcal{O}$ denotes big-O and $\mathcal{\tilde{O}}$ is soft-O. $\delta_{l,F}$, $\delta_{l,2}$, and $\delta_{l,2 \to 1}$ denotes upper bounds of the Frobenius norm $\| \omega_l \|_F \leq \delta_{l,F}$, spectral norm $\| \omega_l \|_2 \leq \delta_{l,2}$ and the sum of the Euclidean norms for all rows $\| \omega_l \|_{2 \to 1} \leq \delta_{l,2 \to 1}$ ($\ell_{2 \to 1}$) of weights $\omega_l$ at the $l^{th}$ layer of an $L$ layer DNN using $N$ samples. Suppose that all layers have the same width $\varpi$, weights have the same length $\mathcal{K}$ and the same stride $\mathfrak{s}$. Then, generalization bounds are obtained for DNNs using these fixed parameters by $\| \omega_l \|_2 = \frac{\mathcal{K}}{\mathfrak{s}}$, $\| \omega_l \|_F = \sqrt{\varpi}$ and $\| \omega_l \|_{2 \to 1} = \varpi$. We compute a concatenated weight matrix $\omega_{g,l} = (\omega_{g,l}^1, \omega_{g,l}^2, \ldots, \omega_{g,l}^{|\mathfrak{g}|})$ for the $g^{th}$ weight group of size $|\mathfrak{g}|, g=1,2,\ldots,G_l, \forall l$  using a weight grouping strategy.  Then, we have upper bounds of norms by $ \| \omega_{g,l}\|_F \leq \delta_{g,l,F} \leq 1$, $\| \omega_{g,l}\|_2 \leq\delta_{g,l,2}  \leq 1$ and $\| \omega_{g,l}\|_{2 \to 1} \leq \delta_{g,l,2 \to 1}  \leq 1, g=1,2,\ldots,G_l$, which are defined in Table~\ref{tab:norms}. }	
	\begin{tabular}{|C{3.25cm}|C{5.6cm}|}
		\toprule
		\toprule
		& \multicolumn{1}{|c|}{\textbf{DNNs} (dynamic group scaling)}\\
		\bottomrule
		Neyshabur et al.~\cite{Neyshabur15}     & $\mathcal{O}\Big( \frac{2^L \prod\limits_{l=1}^{L} \prod\limits_{g=1}^{G_l} \delta_{g,l,F}}{\sqrt{N}} \Big)$ \\
		Bartlett et~al.~\cite{NIPS2017204}   & $\mathcal{\tilde{O}} \Bigg( \frac{\prod\limits _{l=1} ^L \prod\limits _{g=1} ^{G_l} \delta_{g,l,2}}{\sqrt{N}} \Big( \sum \limits_{l=1} ^L \prod \limits _{g=1}^{G_l} (\frac{\delta_{g,l,2 \to 1}}{\delta_{g,l,2}})^{\frac{2}{3}} \Big) ^{\frac{3}{2}} \Bigg)$\\
		Neyshabur et al.~\cite{neyshabur2018a} & $\mathcal{\tilde{O}} \Bigg ( \frac{\prod \limits_{l=1}^{L} \prod \limits_{g=1}^{G_l}\delta_{g,l,2}} {\sqrt{N}} \sqrt{L^2 \varpi \sum\limits_{l=1} ^L \prod \limits_{g=1}^{G_l}\frac{\delta^2_{g,l,F}}{\delta^2_{g,l,2} } } \Bigg)$ \\
		\bottomrule
		\bottomrule
		
	\end{tabular}%
	\label{tab:compareGen}%
\end{table*}%

\section{Proofs of Theorems given in the Main Text}

\begin{definition}[Sectional curvature of component manifolds]
	Let $\mathfrak{X}(\Mi)$ denote the set of smooth vector fields on $\Mi$. The sectional curvature of $\Mi$ associated with a two dimensional subspace $\mathfrak{T} \subset \mathcal{T}_{\Oi}\Mi$  is defined by 
	\begin{equation}
	\mathfrak{c}_{\iota} = \frac{\left\langle \Ci(X_{\Oi},Y_{\Oi})Y_{\Oi},X_{\Oi}  \right\rangle}{\left\langle  X_{\Oi} , X_{\Oi} \right\rangle \left\langle  Y_{\Oi} , Y_{\Oi} \right\rangle  -   \left\langle X_{\Oi} ,Y_{\Oi} \right\rangle^2}
	\end{equation}
	where $\Ci(X_{\Oi},Y_{\Oi})Y_{\Oi}$ is the Riemannian curvature tensor, $\left\langle \cdot,\cdot \right\rangle$ is an inner product, ${X_{\Oi} \in \mathfrak{X}(\Mi)}$ and ${Y_{\Oi} \in \mathfrak{X}(\Mi)}$ form a basis of $\mathfrak{T}$.\QEDbs	
\end{definition}

\begin{definition}[Riemannian connection on component embedded weight manifolds]
	Let $\mathfrak{X}(\Mi)$ denote the set of smooth vector fields on $\Mi$ and $\mathfrak{F}(\Mi)$ denote the set of smooth scalar fields on $\Mi$.The Riemannian connection $\bar{\nabla}$ on $\Mi$ is a mapping \cite{absil_retr}
	\begin{equation}
	\bar{\nabla}: \mathfrak{X}(\Mi) \times \mathfrak{X}(\Mi) \to \mathfrak{X}(\Mi): (X_{\Oi}, Y_{\Oi} ) \mapsto \bar{\nabla} X_{\Oi}Y_{\Oi}
	\end{equation}
	which satisfies the following properties:
	\begin{enumerate}
		\item $\bar{\nabla}_{pX_{\Oi}+qY_{\Oi}} Z_{\Oi} = p\bar{\nabla}_{Z_{\Oi}} + q \nabla_{Y_{\Oi}} Z_{\Oi}$,
		\item $\bar{\nabla} X_{\Oi}(\alpha Y_{\Oi} + \beta Z_{\Oi}) = \alpha \bar{\nabla}_{X_{\Oi}Y_{\Oi}} + \beta \bar{\nabla}_{X_{\Oi}}Z_{\Oi}$,
		\item $ \bar{\nabla} _{X_{\Oi}}(pY_{\Oi} ) = (X_{\Oi}p)Y_{\Oi} + p\bar{\nabla}_{X_{\Oi}}Y_{\Oi}$,
		\item $\bar{\nabla}_{X_{\Oi}} Y_{\Oi} - \bar{\nabla}_{Y_{\Oi}} X_{\Oi} = [X_{\Oi},Y_{\Oi}] $ and 
		\item $Z_{\Oi} \left\langle  X_{\Oi},Y_{\Oi} \right\rangle  = \left\langle \bar{\nabla}_{Z_{\Oi}}X_{\Oi}, Y_{\Oi}\right\rangle + \left\langle X_{\Oi}, \bar{\nabla}_Z Y_{\Oi} \right\rangle$
	\end{enumerate}
	where $X_{\Oi}, Y_{\Oi}, Z_{\Oi} \in \mathfrak{X}(\Mi)$, $p, q \in \mathfrak{F}(\Mi)$, $\alpha, \beta \in \mathbb{R}$, $\left\langle \cdot,\cdot \right\rangle$ is an inner product, $[X_{\Oi},Y_{\Oi}]$ is the Lie bracket of $X_{\Oi}$ and $Y_{\Oi}$, and defined by ${[X_{\Oi}, Y_{\Oi}]p = X_{\Oi}(Y_{\Oi}p) - Y_{\Oi}(X_{\Oi}p)}$, $\forall p \in \mathfrak{F}(\Mi)$. 
	
\end{definition}

\begin{customlemma}{1}[Metric and curvature properties of POMs]
	\label{lemma11}
	Suppose that	$u_{\iota} \in \mathcal{T}_{\omega_{\iota}} \mathcal{M}_{\iota}$ and $v_{\iota} \in \mathcal{T}_{\omega_{\iota}} \mathcal{M}_{\iota}$ are tangent vectors belonging to the tangent space $\mathcal{T}_{\omega_{\iota}} \mathcal{M}_{\iota}$ computed at ${{\omega_{\iota}} \in \mathcal{M}_{\iota}}$, $\forall \iota \in \mathcal{I}_{{G}_l}$. Then, tangent vectors $u_{G_l} \in \mathcal{T}_{\omega_{G_l}} \mathbb{M}_{G_l}$ and $v_{G_l} \in \mathcal{T}_{\omega_{G_l}} \mathbb{M}_{G_l}$ are computed at $\omega_{G_l} \in \mathbb{M}_{G_l}$ by concatenation as $u_{G_l} = (u_1, u_2, \cdots, u_{|\mathcal{I}_{{G}_l}|})$ and 
	$v_{G_l} = (v_1, v_2, \cdots, v_{|\mathcal{I}_{{G}_l}|})$. If each weight manifold $\mathcal{M}_{\iota}$ is endowed with a Riemannian metric $\mathfrak{d}_{\iota}$, then a $G_l$-POM is endowed with the metric $\mathfrak{d}_{G_l}$ computed by
	\begin{equation}
	\mathfrak{d}_{G_l} ( u_{G_l} , v_{G_l} ) = \sum \limits _{\iota \in \mathcal{I}_{{G}_l}} \mathfrak{d}_{\iota}(u_{\iota},v_{\iota}).
	\label{eq:prod_metric}
	\end{equation}
	In addition, suppose that $\bar{C}_{\iota}$ is the Riemannian curvature tensor field (endomorphism) \cite{lee2009manifolds} of $\mathcal{M}_{\iota}$,  ${x_{\iota}, y_{\iota} \in \mathcal{T}_{\omega_{\iota}} \mathcal{M}_{\iota}}$, $\forall \iota \in \mathcal{I}_{{G}_l}$ defined by
	\begin{equation}
	\bar{C}_{\iota}(u_{\iota},v_{\iota},x_{\iota},y_{\iota}) = \left\langle {C}_{\iota} (U,V)X,Y \right\rangle_{\Oi}, 
	\label{eq:R_tensor}
	\end{equation}
	where $U,V,X,Y$ are vector fields such that $U_{\Oi} = u_{\iota} $, $V_{\Oi} = v_{\iota} $, $X_{\Oi} = x_{\iota} $, and $Y_{\Oi} = y_{\iota} $. Then, the Riemannian curvature tensor field $\bar{C}_{G_l} $ of $\mathbb{M}_{G_l}$ is computed by
	\begin{equation}
	\bar{C}_{G_l} ( u_{G_l} , v_{G_l}, x_{G_l} , y_{G_l}  ) = \sum \limits _{\iota \in \mathcal{I}_{{G}_l}} \bar{C}_{\iota}(u_{\iota},v_{\iota},x_{\iota},y_{\iota}),
	\label{eq:curv_tensor}
	\end{equation}
	where  ${x_{G_l} = (x_1, x_2, \cdots, x_{|\mathcal{I}_{{G}_l}|})}$ and 
	$y_{G_l} = (y_1, y_2, \cdots, y_{|\mathcal{I}_{{G}_l}|})$. 
	Moreover, $\mathbb{M}_{G_l}$ has never strictly positive sectional curvature $\mathfrak{c}_{G_l}$ in the metric \eqref{eq:prod_metric}. In addition, if $\mathbb{M}_{G_l}$ is compact, then $\mathbb{M}_{G_l}$ does not admit a metric with negative sectional curvature $\mathfrak{c}_{G_l}$. \QEDbs
	
\end{customlemma}

\begin{proof}
	
	Since each weight manifold $\Mi$ is a Riemannian manifold, $\mathfrak{d}_{\iota}$ is a Riemannian metric such that ${\mathfrak{d}_{\iota}(u_{\iota}, v_{\iota}) = \left\langle u_{\iota}, v_{\iota} \right\rangle }$. Thereby,
	\begin{equation}
	\mathfrak{d}_{G_l} ( u_{G_l} , v_{G_l} ) =  \left\langle u_{G_l}, v_{G_l} \right\rangle 
	= \sum \limits _{\iota \in \mathcal{I}_{{G}_l}} \left\langle u_{\iota}, v_{\iota} \right\rangle 	
	\sum \limits _{\iota \in \mathcal{I}_{{G}_l}} \mathfrak{d}_{\iota}(u_{\iota},v_{\iota})
	\label{eq:prod_metric2}
	\end{equation}
	and we obtain \eqref{eq:prod_metric}. In order to derive \eqref{eq:curv_tensor}, we first compute
	\begin{equation}	
	\left\langle \sum \limits _{\iota \in \mathcal{I}_{{G}_l}} u_{\iota}, \sum \limits _{\iota \in \mathcal{I}_{{G}_l}} v_{\iota} \right\rangle  = \sum \limits _{\iota \in \mathcal{I}_{{G}_l}} \left\langle  u_{\iota}, v_{\iota} \right\rangle .
	\label{eq:rm1}
	\end{equation}  
	Then, we use the equations for the Lie bracket by
	\begin{equation}	
	\left[ \sum \limits _{\iota \in \mathcal{I}_{{G}_l}} u_{\iota}, \sum \limits _{\iota \in \mathcal{I}_{{G}_l}} v_{\iota} \right ] = \sum \limits _{\iota \in \mathcal{I}_{{G}_l}} \left[  u_{\iota}, v_{\iota} \right] .
	\label{eq:rm2}
	\end{equation}
	Next, we employ the Koszul's formula \cite{lee2009manifolds} by
	\begin{equation}
	2 \left\langle \bar{\nabla}_{u_{\iota}} v_{\iota} , x_{\iota} \right\rangle  =
	u_{\iota} \left\langle v_{\iota} , x_{\iota} \right\rangle + v_{\iota}  \left\langle x_{\iota} , u_{\iota} \right\rangle \nonumber 
	- x_{\iota} \left\langle u_{\iota} , v_{\iota} \right\rangle  + \left\langle  x_{\iota} , [u_{\iota} , v_{\iota} ]\right\rangle \nonumber 
	- \left\langle  v_{\iota} , [u_{\iota} , x_{\iota} ]\right\rangle - \left\langle  u_{\iota} , [v_{\iota} , x_{\iota} ]\right\rangle
	\end{equation}	
	such that 
	\begin{equation}	
	\bar{\nabla}_{ \bar{u}}
	( \bar{v} ) = \sum \limits _{\iota \in \mathcal{I}_{{G}_l}} \bar{\nabla}_{u_{\iota}} (v_{\iota} ),
	\label{eq:rm3}
	\end{equation}
	where $\bar{u} = \sum \limits _{\iota \in \mathcal{I}_{{G}_l}} u_{\iota} $ and $\bar{v} = \sum \limits _{\iota \in \mathcal{I}_{{G}_l}} v_{\iota} $. Using \eqref{eq:R_tensor} and definition of the curvature with \eqref{eq:prod_metric2}, \eqref{eq:rm1}, \eqref{eq:rm2}, and \eqref{eq:rm3}, we obtain \eqref{eq:curv_tensor}. 
	
	In order to show that $\mathbb{M}_{G_l}$ has never strictly positive sectional curvature $\mathfrak{c}_{G_l}$ in the metric \eqref{eq:prod_metric}, it is sufficient to show that some sectional curvatures always vanish. Suppose that $U$ is a vector field on $\mathbb{M}_{G_l}$ along a component weight manifold $\Mi$ such that  no local coordinate $o$ of $\mathcal{M}_{\bar{\iota}}$ and $\frac{\partial}{\partial o}$ are present in local coordinates of $U$, $\forall \iota \neq \bar{\iota}$, $\bar{\iota} \in \mathcal{I}_{G_l}$.
	In addition, suppose that $\bar{U}$ is a vector field along $\mathcal{M}_{\bar{\iota}}$. Then, $\bar{\nabla}_{U} \bar{U} = 0$, $\forall \iota, \bar{\iota} \in \mathcal{I}_{G_l}$. By employing \eqref{eq:rm3}, we have $\bar{C}_{\iota}(u_{\iota},v_{\iota},x_{\iota},y_{\iota}) = 0$. Then, we use \eqref{eq:curv_tensor} to obtain $\bar{C}_{G_l} ( u_{G_l} , v_{G_l}, x_{G_l} , y_{G_l}  ) = 0$. Therefore, following the definition of the sectional curvature, for arbitrary vector fields on component manifolds, $\mathbb{M}_{G_l}$ has never strictly positive sectional curvature $\mathfrak{c}_{G_l}$ in the metric \eqref{eq:prod_metric}. Since $\mathbb{M}_{G_l}$ is a Riemannian manifold, if $\mathbb{M}_{G_l}$ is compact, then $\mathbb{M}_{G_l}$ does not admit a metric with negative sectional curvature $\mathfrak{c}_{G_l}$ by the Preissmann's theorem \cite{petersen2006riemannian} \footnote{see Theorem 24 in \cite{petersen2006riemannian} .}.

\end{proof}

\begin{customthm}{1}[Computation of gradients on tangent spaces]
	\label{thm_grads}
	The $\ell_2$ norm $\| \gr \mathcal{L}(\omega_{G^m_l}^{t})  \|_2$ of the gradient $\gr \mathcal{L}(\omega_{G^m_l}^{t})$ residing on  $\mathcal{T}_{\omega^t_{G^m_l}} \mathbb{M}_{G^m_l}$ at the $t^{th}$ epoch and the $l^{th}$ layer can be computed by		
	\vspace{-0.5cm}
	\begin{equation}
	\| \gr \mathcal{L}(\omega_{G^m_l}^{t})  \|_2 = \Big (\sum \limits_{\iota \in \mathcal{I}_{G^m_l}} \gr \mathcal{L}(\omega_{l,\iota}^{t})^2 \Big)^{\frac{1}{2}},
	\vspace{-0.1cm}
	\label{eq:grad_norm}
	\end{equation} 	
	where $\gr \mathcal{L}(\omega_{l,\iota}^{t})$ is the gradient computed for the weight $\omega_{l,\iota}^{t}$  on the tangent space $\mathcal{T}_{\omega^t_{\iota,l}} \mathbb{M}_{\iota}$, ${\forall \iota \in \mathcal{I}_{G^m_l}}$.
	\QEDbs
\end{customthm}

\begin{proof}
	We use the inner product for the Riemannian metric $\mathfrak{d}_{G_l} ( \gr \mathcal{L}(\omega_{G^m_l}^{t}) , \gr \mathcal{L}(\omega_{G^m_l}^{t}) )$ and 
	
	$\mathfrak{d}_{\iota} ( \gr \mathcal{L}(\omega_{l,\iota}^{t}) , \gr \mathcal{L}(\omega_{l,\iota}^{t}) )$ of manifolds $ \mathbb{M}_{G^m_l}$ and $ \mathbb{M}_{\iota}, \forall \iota$, respectively. By definition of the product manifold, we have
	\begin{eqnarray}
	\gr \mathcal{L}(\omega_{G^m_l}^{t}) = \Big ( \gr \mathcal{L}(\omega_{l,1}^{t}), \gr \mathcal{L}(\omega_{l,2}^{t}), 
	\gr \mathcal{L}(\omega_{l,|\mathcal{I}_{G_l}|}^{t}) \Big ).
	\end{eqnarray}
	Thereby, we can apply bilinearity of inner product in Lemma~1 and obtain 
	\begin{equation}
	\| \gr \mathcal{L}(\omega_{G^m_l}^{t})  \|_2^2 = \Big (\sum \limits_{\iota \in \mathcal{I}_{G^m_l}} \gr \mathcal{L}(\omega_{l,\iota}^{t})^2 \Big),
	\label{eqtemp}
	\end{equation}
	where $\| \cdot \|_2^2$ is the squared $\ell_2$ norm. The result follows by applying the square root to \eqref{eqtemp}.
\end{proof}

\begin{customthm}{2}[Convergence of the FG-SGD]
	\label{thm33}
	Suppose that there exists a local minimum ${\hat{\omega}_{G_l} \in \mathbb{M}_{G_l}}, \forall {G_l \subseteq \mathcal{G}_l}$, $\forall l$, and $\exists \epsilon>0$ such that $\inf \limits _{\rho_{G_l}^{t} > \epsilon^{\frac{1}{2}}} \left\langle \phi_{\omega_{G_l}^t}(\hat{\omega}_{G_l})^{-1}, \nabla \mathcal{L}(\omega_{G_l}^t) \right\rangle <0$, where $\phi$ is an exponential map or a twice continuously differentiable retraction, and $\langle \cdot,\cdot \rangle$ is the inner product. Then, the loss function and the gradient converges almost surely (a.s.) by $\mathcal{L}(\omega^t_{G_l}) \xrightarrow[t \to \infty]{\rm a.s.} \mathcal{L}(\hat{\omega}_{G_l})$, and $\nabla \mathcal{L}(\omega^t_{G_l}) \xrightarrow[t \to \infty]{\rm a.s.} 0$, for each $\mathbb{M}_{G_l}, \forall l$. \QEDbs
	
\end{customthm}

\begin{proof}
	In this theorem, we generalize the proof idea of Theorem 4.1 and 4.2 given in \cite{ooAAAI18}, and Theorem 3 given in \cite{sgdman} for collections of products of embedded weight manifolds	(POMs) for training of CNNs. The proof idea is to show that $\rho_{G_l}^{t} \triangleq \rho (\omega_{l,\iota}^{t},\hat{\omega}_{l,\iota})$ converges almost surely to $0$ as $t \to \infty$. For this purpose, we need to first model the change of gradient on the geodesic $\rho_{G_l}^{t}$ by defining a function $\Psi_t \triangleq \psi((\rho_{G_l}^{t})^2)$ according to the following constraints \cite{sgdman};
	\begin{itemize}
		\item $\Psi_t = 0$, for $0 \leq \rho_{G_l}^{t} \leq \sqrt{\epsilon}$.
		\item $0 < \Psi''_t \leq 2$, for $\sqrt{\epsilon} \leq \rho_{G_l}^{t} \leq \sqrt{\epsilon+1}$.
		\item $\Psi'_t = 1$, for $\rho_{G_l}^{t} \geq \sqrt{\epsilon+1}$.
	\end{itemize}  
	Then, we compute gradients and geodesics on collections of POMs using \eqref{eq:prod_metric} given in Lemma~\ref{lemma11} by
	\begin{equation}
	\| \gr \mathcal{L}(\omega_{G_l}^{t})  \|_2 = \Big (\sum \limits_{\omega_{l,\iota}^{t} \in \mathbb{M}_{\iota}, \iota \in \mathcal{I}_{G_l}} \gr \mathcal{L}(\omega_{l,\iota}^{t})^2 \Big)^{\frac{1}{2}}
	\end{equation}
	and
	\begin{equation}
	\rho (\omega_{G_l}^{t})  = \Big (\sum \limits_{\omega_{l,\iota}^{t} \in \mathbb{M}_{\iota}, \iota \in \mathcal{I}_{G_l}} \rho (\omega_{l,\iota}^{t},\hat{\omega}_{l,\iota}) \Big),
	\end{equation}
	where ${\omega^t_{G_l} = (\omega^t_1, \omega^t_2, \cdots, \omega^t_{|\mathcal{I}_{{G}_l}|})}$.
	We employ a Taylor expansion on $\Psi_t$ \cite{sgdman,ooAAAI18}, and we obtain
	\begin{equation}
	\Psi_{t+1} - \Psi_t \leq ((\rho_{G_l}^{t+1})^2 - (\rho_{G_l}^{t})^2 ) \Psi'_t + ((\rho_{G_l}^{t+1})^2 - (\rho_{G_l}^{t})^2 ) ^2.
	\end{equation}
	
	In order to compute the difference between $\rho_{G_l}^{t+1}$ and $\rho_{G_l}^{t}$, we employ a Taylor expansion on the geodesics \cite{sgdman,ooAAAI18} by
	\begin{equation}
	\rho_{G_l}^{t+1} - \rho_{G_l}^{t} \leq  \Big  (\frac{g(t,\Theta)}{\mathfrak{g}(\omega_{G_l}^t)} \Big)^2 \| \gr \mathcal{L}(\omega_{G_l}^{t}) \|^2 \kappa \nonumber 
	- 2 \left\langle h(\gr \mathcal{L}(\omega_{G_l}^{t}), g(t,\Theta)) , \phi_{\omega_{G_l}^t}(\hat{\omega}_{G_l})^{-1} \right\rangle ,
	\label{eq:geod_diff}
	\end{equation}
	where ${\hat{\omega}_{G_l} = (\hat{\omega}_1, \hat{\omega}_2, \cdots, \hat{\omega}_{|\mathcal{I}_{{G}_l}|})}$, and $\kappa \leq \Upsilon_1$ where $\Upsilon_1=1+\mathfrak{c}_{G_l}(\rho_{G_l}^{t} + R_{G_l}^{t})$ is an upper bound on the operator norm of half of the Riemannian Hessian of $\rho(\cdot,\hat{\omega}_{G_l})^2$ along the geodesic joining $\omega_{G_l}^{t}$ and $\omega_{G_l}^{t+1}$. In order to explore asymptotic convergence, we define $\Omega_t = \{s_i \}_{i=1}^{t-1}$ to be an increasing sequence of $\sigma$ algebras generated by samples that are processed before the $t^{th}$ epoch. Since $s_t$ is independent of $\Omega_t$ and  $\omega_{G_l}^t$ is $\Omega_t$  measurable, we have 
	\begin{equation}
	\mathbb{E} (h(\gr \mathcal{L}(\omega_{G_l}^{t}), g(t,\Theta))^2 \kappa | \Omega_t] ) \leq 
	\Big  (\frac{g(t,\Theta)}{\mathfrak{g}(\omega_{G_l}^t)} \Big)^2 \mathbb{E} \Big ( (R_{G_l}^{t} )^2 \Upsilon_1 \Big), 
	\end{equation} 
	and
	\begin{equation}
	\mathbb{E}((\rho_{G_l}^{t+1})^2 - (\rho_{G_l}^{t})^2 |\Omega_t )  \leq 2 \frac{g(t,\Theta)}{\mathfrak{g}(\omega_{G_l}^t)} \left\langle \phi_{\omega_{G_l}^t}(\hat{\omega}_{G_l})^{-1}, \nabla \mathcal{L}(\omega_{G_l}^t) \right\rangle + g(t,\Theta)^2.
	\end{equation}
	If $\mathfrak{g}(\omega_{G_l}^t) = \max\{ 1,\Gamma_1^t\}^{\frac{1}{2}}$, $\Gamma_1^t = (R_{G_l}^{t})^2 \Gamma_2^t$, ${\Gamma_2^t = \max \{(2\rho_{G_l}^{t} + R_{G_l}^{t})^2, (1+\mathfrak{c}_{G_l}(\rho_{G_l}^{t} + R_{G_l}^{t}))\} }$, then we have
	\begin{equation}
	\mathbb{E} (\Psi_{t+1} - \Psi_t| \Omega_t) \leq \mathbb{E}((\rho_{G_l}^{t+1})^2 - (\rho_{G_l}^{t})^2 |\Omega_t ) \Psi'_t +  g(t, \Theta ) ^2
	\end{equation}
	and
	\begin{equation}
	\mathbb{E} (\Psi_{t+1} - \Psi_t| \Omega_t) \leq 2 \frac{g(t,\Theta)}{\mathfrak{g}(\omega_{G_l}^t)} \left\langle \phi_{\omega_{G_l}^t}(\hat{\omega}_{G_l})^{-1}, \nabla \mathcal{L}(\omega_{G_l}^t) \right\rangle  \Psi'_t +  g(t, \Theta ) ^2.
	\end{equation}
	Thus, we have 
	\begin{equation}
	\mathbb{E} (\Psi_{t+1} - \Psi_t| \Omega_t) \leq 2 g(t, \Theta ) ^2,
	\end{equation}
	and $\Psi_t +\sum_{t=0} ^{\infty} g(t,\Theta)^2$ is a positive supermartingale, and converges almost surely. Since 
	\begin{equation}
	\sum_{t=0} ^{\infty}  \mathbb{E} ( [\mathbb{E} (\Psi_{t+1} - \Psi_t| \Omega_t) ^+ ] )\leq \sum_{t=0} ^{\infty} g(t,\Theta)^2 < \infty,
	\end{equation}
	we observe that $\Psi_t$ is a quasi-martingale \cite{sgdman,ooAAAI18}, and thereby we have almost surely
	\begin{equation}
	- \sum_{t=0} ^{\infty} \frac{g(t,\Theta)}{\mathfrak{g}(\omega_{G_l}^t)} \left\langle \phi_{\omega_{G_l}^t}(\hat{\omega}_{G_l})^{-1}, \nabla \mathcal{L}(\omega_{G_l}^t) \right\rangle \Psi'_t < \infty.
	\end{equation}
	Using properties of quasi-martingale \cite{fisk},  $\Psi_t$ converges almost surely. In order to show almost sure convergence of $\nabla \mathcal{L}(\omega^t_{G_l}) $ to $0$, we use Theorem 4.1 and 4.2 of \cite{ooAAAI18}. For this purpose, we need to show that gradients of loss functions are bounded in compact sets of weights. Since 
	\[\inf \limits _{\rho_{G_l}^{t} > \epsilon^{\frac{1}{2}}} \left\langle \phi_{\omega_{G_l}^t}(\hat{\omega}_{G_l})^{-1}, \nabla \mathcal{L}(\omega_{G_l}^t) \right\rangle <0, 
	\]
	a weight $\omega_{G_l}^t$ is moved towards $\hat{\omega}_{G_l}$ by the gradient when $\rho_{G_l}^{t} > \epsilon^{\frac{1}{2}}$ where the set  $\mathfrak{S}=\{\omega_{G_l}^t : \rho_{G_l}^{t} \leq \epsilon^{\frac{1}{2}} \}$ is a compact set. Since all continuous functions of $\omega_{G_l}^t  \in \mathfrak{S}$ are bounded, and adaptive step size $\mathfrak{g}(\omega_{G_l}^t)$ satisfies $\frac{g(t,\Theta)}{\mathfrak{g}(\omega_{G_l}^t)} \leq g(t,\Theta)$ and $\mathfrak{g}(\omega_{G_l}^t)^2$ dominates $R_{G_l}^t$, we obtain that $\mathbb{E}( R_{G_l}^t)^2 \leq \mathfrak{K}$ for some $\mathfrak{K} > 0$ on a compact set $\mathcal{K}$. Thereby, we can show that conditions of Theorem 4.1 and 4.2 of \cite{ooAAAI18} are satisfied. Therefore, we obtain almost sure convergence of $\nabla \mathcal{L}(\omega^t_{G_l}) $ to $0$ by applying Theorem 4.1 and 4.2 in the rest of the proof.	 
	
\end{proof}

\begin{customcorr}{1}
	\label{corr1}
	Suppose a DNN has loss functions whose local minima are also global minima. If the DNN is trained using the proposed FG-SGD and weight renormalization methods, then the loss of the DNN converges to global minima.	
\end{customcorr}

\begin{proof}
	By Theorem~2, we assure that a loss function of a DNN which employs the proposed FG-SGD and weight renormalization methods for training converges to local minima. If the local minima is the global minima for the DNN, then the loss function converges to the global minima.
\end{proof}

\begin{customcorr}{2}
	\label{corr34}
	Suppose that $\mathbb{M}_{\iota}$ are identified by ${n_{\iota} \geq 2}$ dimensional unit sphere $\mathbb{S}^{n_{\iota}}$, and $\rho_{G_l}^t \leq \hat{\mathfrak{c}}^{-1}$, where $\hat{\mathfrak{c}}$ is an upper bound on the sectional curvatures of $\mathbb{M}_{G_l}, \forall l$ at $\omega_{G_l}^t \in \mathbb{M}_{G_l}, \forall t$. If step size is computed using 
	\begin{equation}
	h(\gr \mathcal{L}(\omega_{G_l}^{t}), g(t,\Theta)) = -\frac{g(t,\Theta)}{\mathfrak{g}(\omega_{G_l}^t)}\gr \mathcal{L}(\omega_{G_l}^{t}),
	\label{eq:steps}
	\end{equation}
	with ${\mathfrak{g}(\omega_{G_l}^t) = (\max\{ 1, (R_{G_l}^{t})^2(2+R_{G_l}^{t})^2 \} })^{\frac{1}{2}}$, then ${\mathcal{L}(\omega^t_{G_l}) \xrightarrow[t \to \infty]{\rm a.s.} \mathcal{L}(\hat{\omega}_{G_l})}$, and ${\nabla \mathcal{L}(\omega^t_{G_l}) \xrightarrow[t \to \infty]{\rm a.s.} 0}$, for each $\mathbb{M}_{G_l}, \forall l$. \QEDbs

\end{customcorr}

\begin{proof}
	If $\mathbb{M}_{G_l}$ is a product of ${n_{\iota} \geq 2}$ dimensional unit spheres $\mathbb{S}^{n_{\iota}}$, then $\mathfrak{c}_{G_l} =0$ and $\hat{\mathfrak{c}}=1$ by Lemma~\ref{lemma11}. Thereby, Theorem~\ref{thm33} is applied to assure convergence by $\Gamma^1_t = (R_{G_l}^{t})^2(2+R_{G_l}^{t})^2$. 
	
\end{proof}

\section{Experimental Details}

We use three benchmark image classification datasets, namely Cifar-10, Cifar-100 and Imagenet \cite{Alexnet}, for analysis of convergence properties and performance of CNNs trained using FG-SGD. The Cifar-10 dataset consists of 60000 $32 \times 32$ RGB images (50000 training images and 10000 test images) in 10 classes, with 6000 images per class. The Cifar-100 dataset consists of 100 classes containing 600 images each (500 training images and 100 testing images per class). The Imagenet (ILSVRC 2012) dataset   consists   of   1000 classes  of  $224 \times 224$ RGB images (1.2  million training images, 100000 test images and 50000 images used for validation).

\subsection{Computational Complexity of Algorithm~1} 
\label{sec:comp}

Compared to SGD algorithms that use weights belonging to linear weight spaces \cite{res_net,nature_deep}, the computational complexity of Algorithm~1 is dominated by computation of the maps $\Pi$ and $\phi$ at line 6 and 9, depending on the structure of the weight manifold used at the $l^{th}$ layer. Concisely, the computational complexity of $\Pi$ is determined by computation of different norms that identify the manifolds. For instance, for the sphere, we use ${\Pi_{\omega_l^t} \mu_t \triangleq (1- \| \omega_l^t \|_F^2) \mu_t}$. Thereby, for an $A \times A$ weight, the complexity is bounded by $O(A^3)$, where $O(\cdot)$ denotes an asymptotic upper bound \cite{algo}. Similarly, the computational complexity of $\phi$ depends on the manifold structure. For example, the exponential maps on the sphere and the oblique manifold can be computed using functions of $\sin$ and $\cos$ functions, while that on the Stiefel manifold is a function of matrix exponential. For computation of matrix exponential, various numerical approximations with ${O}(\epsilon A^3)$ complexity were proposed for different approximation order $\epsilon$ \cite{fisk,Higham,kenney,nineteen}. However, unit norm matrix normalization is used for computation of retractions on the sphere and the oblique manifold. Moreover, QR decomposition of matrices is computed with ${O}(A^3)$ \cite{golub} for retractions on the Stiefel manifold. In addition, computation time of maps can be reduced using parallel computation methods. For instance, a rotation method was suggested to compute QR using ${O}(A^2)$ processors in ${O}(A)$ unit time in \cite{fast_qr}. Therefore, computation of retractions is computationally less complex compared to that of the exponential maps. Since the complexity analysis of these maps is beyond the scope of this work, and they provide the same convergence properties for our proposed algorithm, we used the retractions in the experiments. Implementation details are given in the next section.

\subsubsection{A Discussion on Implementation of Algorithm~1 in Parallel and Distributed Computing Systems} 		

In the experiments, 
algorithms are implemented using GPU and CPU servers consisting of GTX 2070, GTX 1080, GTX-Titan-X, GTX-Titan-Black, Intel i7-5930K, Intel Xeon E5-1650 v3 and  E5-2697 v2. Since we used hybrid GPU and CPU servers in the experiments, and a detailed analysis of parallel and distributed computation methods of CNNs is beyond the scope of this work, we report bounds on average running times of SGD algorithms in this section.

In the implementation of linear Euclidean SGD methods, we use vectorized computation of weight updates. Therefore, we use large scale matrix computation methods (in some cases, for sparse matrices)  to improve running time of the linear Euclidean SGD methods. However, we deal with optimization using batched (small size) dense matrices in the implementation of Algorithm~1 \cite{964}. Therefore, in order to improve running time of the algorithm, we implemented Algorithm~1 using hybrid CPU-GPU programming paradigms. 

More precisely, we consider two computation schemes according to matrix/tensor structure of the weights, i.e. geometric structure of weight manifolds. First, we recall that we construct different manifolds of weights ${ \mathcal{W} = \{ \mathbf{W}_{d,l} \in \mathbb{R}^{A_l \times B_l \times C_l} \} _{d=1} ^{D_l}}, \forall l=1,2,\dots,L$, at different layers of an $L$-layer CNN. Then, we implement projections of gradients and retractions at
\begin{enumerate}
	
	\item Fully Connected (FC) layers at which we use ${\mathbf{W}_{l}^{fc} \in \mathbb{R}^{C_l \times D_l}}$ with $A_l = B_l =1$, and
	
	\item Convolution (Conv) layers at which we use $\mathbf{W}_{d,l} \in \mathcal{W}$ with $A_l > 1$ and $B_l >1$.
	
\end{enumerate}

At the FC layers, we implemented Algorithm~1 on GPUs using Cuda with Cublas and Magma \cite{dghklty14,tdb10,tnld10} Blas \cite{ntd10_vecpar,ntd10}. In the experimental analyses, we obtained similar running times using Cublas and Magma Blas implementation of Algorithm~1 (denoted by $\mathcal{R}^{fc}_M$) compared to running time of linear Euclidean SGD (denoted by $\mathcal{R}^{fc}_E$), for each epoch. 

For instance, if we train CNNs using the Cifar-100 dataset and one GTX 1080, then we observe $\mathcal{R}^{fc}_M < \mathfrak{a} \mathcal{R}^{fc}_E$, where the running times are bounded by $\mathfrak{a} > 0$ due to implementation of gradient projections and retractions. The overhead factor $\mathfrak{a}$ also depends on the manifold structure of the weights such that $\mathfrak{a} < 1.5$ for the sphere, $\mathfrak{a} <2.5$ for the oblique manifold and $\mathfrak{a} < 5$ for the Stiefel manifold. 

When we implemented a QR decomposition algorithm using the Givens transformation (Rotation) \cite{Brouwer2014,golub}, we obtained further improvement by $\mathfrak{a} < 4$. In addition, batch size does not affect the overhead of running time crucially as long as the GPU memory is sufficient. The effect of this overhead on the overall training time depends on structure of CNNs. For example, we use multiple (6) FC layers in NiNs where we have 2 FC layers in SKs. Therefore, the overhead affects the training time of NiNs more than that of SKs. 

At the Conv layers, we implemented Algorithm~1 on both GPUs and CPUs. However, the structure of parallelization of projections and maps at the Conv layers is different than that of projections and maps computed at the FC layers. More precisely, we perform parallel computation either 1) using tensors $\mathbf{W}_{d,l} \in \mathbb{R}^{A_l \times B_l \times C_l}$ for each output $d=1,2,\dots,D_l$, or 2) using matrices ${W}_{c,d,l} \in \mathbb{R}^{A_l \times B_l}$ for each output $d=1,2,\dots,D_l$ and channel $c=1,2,\ldots,C_l$. 

Since there is an I/O bottleneck between transfer of matrices and tensors to/from GPUs from/to CPUs, we used either (1) or (2) according to output size $D_l$, and channel size $C_l$. For instance, if $C_l > D_l$, then we performed computations on GPUs. Otherwise, we implemented the algorithm on multi-core CPUs. 

In average, for an epoch\footnote{For the example of training using the Cifar-100 dataset given above.}, the running time of a GPU implementation of Algorithm~1 for the case (1) denoted by $\mathcal{R}^{1}_{M,gpu}$, and that of linear Euclidean SGD for the case $\mathcal{R}^{1}_{E,gpu}$ are related by $\mathcal{R}^{1}_{E,gpu} < \mathfrak{a} \mathcal{R}^{1}_{M,gpu}$ for $\mathfrak{a} < 3$ for the sphere and $\mathfrak{a} < 3$ for the oblique manifold and $\mathfrak{a} < 6$ for the Stiefel manifold\footnote{For different implementations of QR decomposition on GPUs, we observed $3<\mathfrak{a} < 6.$}. The additional computational overhead can be attributed to additional transmission time and computation of multi-dimensional transpose operations. 

Moreover, we observed that the running time of the multi-core CPU implementation of the algorithm $\mathcal{R}^{1}_{M,cpu}$ is bounded by $\mathcal{R}^{1}_{M,gpu} < \mathfrak{a} \mathcal{R}^{1}_{M,cpu}$ for ${\mathfrak{a} < f(D_l)< 10}$, where $f(\cdot)$ is a function of number of output $D_l$ for all manifolds\footnote{We observed that for Intel Xeon E5-1650 v3, and obtained improvement of running time by approximately $f(D_l)<5$ for E5-2697 v2 since using larger number of CPU cores.}. In other words, the difference between running times on CPUs and GPUs is affected by $D_l$ more than the other parameters $2 \leq A_l \leq 7$ and $2 \leq B_l \leq 7$, and $C_l$. This observation can be attributed to the less overhead between Blas and Cublas implementations of matrix operations for small number (e.g. $C_l<10^3$) of weight matrices. 

For the second case where $C_l>D_l$, we observed that $\mathcal{R}^{1}_{E,gpu} < \mathfrak{a}_1 \mathcal{R}^{1}_{M,cpu} < \mathfrak{a}_2 \mathcal{R}^{1}_{M,gpu}$. We observed that ${\mathfrak{a}_1 < \hat{f}(C_l,D_l) < 2}$ and ${\mathfrak{a}_2 < \hat{f}(C_l,D_l) < 5}$, where $\hat{f}(\cdot,\cdot)$ is a function of both $C_l$ and $D_l$, for the sphere, and scales for the other manifolds accordingly, for implementation using one GTX 1080 and E5-2697 v2. 

\subsection{Implementation Details of Algorithm~1}

In this section, we give implementation details of Algorithm~1.

\subsubsection{Identification of Component Kernel Submanifolds of POMs} 

We identify component weight manifolds $\mathcal{M}_{\iota}$ of POMs $\mathbb{M}_{G_l}$ at each $l^{th}$ of an $L$-layer CNN, and initialize weights residing in the manifolds considering both statistical properties of data, and geometric properties of weight manifolds. 



In the experiments, we used the sphere, the oblique manifold and the Stiefel manifold to construct component weight manifolds   according to definition of manifolds given in Table~\ref{tab:manifolds}. 

\begin{table*}[h]
	\caption{Tangent spaces and maps used for orthogonal projection of Euclidean gradients obtained using backpropagation onto the tangent spaces for the manifolds of the normalized weights defined in Table~\ref{tab:manifolds}. We denote a vector realized by a Euclidean gradient obtained at a weight $\omega^t_{G_l}$ from the $l+1^{st}$ layer using backpropagation by $\mu \triangleq \Big ( \gr_E \; \mathcal{L}(\omega_{g,l}^{t}),\Theta,\mathcal{R}_l^t \Big)$ (see Line 5 of Algorithm~1).}	
	\vspace{0.5cm}
	\begin{minipage}{\textwidth} 
		
		\begin{center}
			
			\begin{tabular}{ccc}
				\hline
				\textbf{Manifolds} & \textbf{Tangent Spaces} & \textbf{Projection of Gradients}  \\
				\hline
				
				\\
				
				$\mathcal{S}(A_l,B_l)$ &	$ T_{\omega} \mathcal{S}(A_l,B_l) = \{ \hat{\omega} \in \mathbb{R}^{A_l \times B_l}: \omega^{\rm T} \hat{\omega} = 0  \}$ & $\Pi_{\omega} \mu = (I-\omega \omega^{\rm T}) \mu$  \\
				
				\\
				
				$\mathcal{OB}(A_l,B_l)$ & $ {T_{\omega} \mathcal{OB}(A_l,B_l) =  \{ \hat{\omega} \in \mathbb{R}^{A_l \times B_l}: {\omega}^{\rm T} \hat{\omega} = 0  \}}$  & $ \Pi_{\omega} \mu = \mu - \omega {\rm ddiag} (\omega^{\rm T} \mu)$ \\ 				
				
				\\
				
				$St(A_l,B_l)$ & $T_{\omega} St(A_l,B_l) = \{ \hat{\omega} \in \mathbb{R}^{A_l \times B_l}: { \rm ddiag} (\omega^{\rm T} \hat{\omega})= 0  \}$ & $\Pi_{\omega} \mu = (I - \omega \omega^{\rm T} ) \mu + \omega \varsigma(\omega^{\rm T} \mu)$ \\		
				
				\\

				\hline

			\end{tabular}
		\end{center}
	\end{minipage}%
	\vspace{0.2cm}
	\label{tab:tangent_spaces}%
\end{table*}%

\begin{table*}[h]
	\caption{Exponential maps and retractions for the manifolds of the normalized weights defined in Table~\ref{tab:manifolds}. We denote a vector moved on a tangent space at the $t^{th}$ epoch by $v_t$ (see Line 8 of Algorithm~1). In addition, $\aleph(Z)$ is the unit-norm normalization of each column of a matrix $Z$. $\mathcal{Q}_{\mathcal{F}}(Z) := Q$ is the $Q$ factor of the QR decomposition $Z =QR$ of $Z$. }	
	\vspace{0.5cm}
	\begin{minipage}{\textwidth} 
		
		\begin{center}
			
			\begin{tabular}{ccc}
				\hline
				\textbf{Manifolds} & \textbf{Exponential Maps} & \textbf{Retraction} \\
				\hline
				\\
				
				$\mathcal{S}(A_l,B_l)$ &	$  \exp_{\omega}(v) ={\omega}\cos(\|v \|_F) +  \frac{v}{\| v \|_F}\sin(\| v \|_F) $  & $\mathfrak{R}_{{\omega}}(v) = \frac{\omega + v}{\| \omega +v  \|_F} $ \\
				\\
				
				$\mathcal{OB}(A_l,B_l)$ & ${ \exp_{\omega} (v) = \omega {\rm ddiag } (\cos(\| v \|_F)) + v {\rm ddiag} ( \frac{\sin(\| v\|_F)}{\| v \|_F} )}$ & $\mathfrak{R}_{{\omega}}(v) = \aleph(\omega+v)$ \\
				
				\\

				$St(A_l,B_l)$ &  $\exp _{\omega} (v) = [ \omega \; v ] \hat{\exp}   \Big ( \begin{bmatrix}
				\omega ^{\rm T} v & - v^{\rm T} v\\
				I & \omega ^{\rm T} v
				\end{bmatrix}  \Big)
				\begin{bmatrix}
				I \\
				0
				\end{bmatrix}     \hat{\exp} (-\omega^{\rm T} v)$ &  $\mathfrak{R}_{{\omega}}(v) = \mathcal{Q}_{\mathcal{F}}(\omega+v)$\\		
				
				\\

				\hline

			\end{tabular}
		\end{center}
	\end{minipage}%
	\vspace{0.5cm}
	
	\label{tab:groups_manifolds}%
\end{table*}%

\subsubsection{Computation of Gradient Maps, Projections and Retractions used in {Algorithm} 1} 
\label{sec:comp}

In this section, we provide the details of the methods used for computation of gradient maps, projections and retractions for different collections of POMs in {Algorithm}~1. We denote a vector moved on a tangent space at the $t^{th}$ epoch by $v_t$ (see Line 7 of Algorithm~1).  In addition, $\aleph(Z)$ is the unit-norm normalization of each column of a matrix $Z$. $\mathcal{Q}_{\mathcal{F}}(Z) := Q$ is the $Q$ factor of the QR decomposition $Z=QR$ of $Z$.

Definitions of component manifolds of POMs used in this work are given in Table~\ref{tab:manifolds}. In Table~\ref{tab:tangent_spaces}, we provide tangent spaces and maps used for orthogonal projection of Euclidean gradients onto the tangent spaces for the manifolds of the normalized weights which are defined in Table~\ref{tab:manifolds}. Exponential maps and retractions are given in Table~\ref{tab:groups_manifolds}.

We also note that various types of projections, exponential maps and retractions can be computed and used in {Algorithm}~1 in addition to the projections, maps and retractions given in the tables. More detailed discussion on their computation are given in \cite{oblq,manopt_book,absil_retr}.


\subsection{Implementation Details of CNN Architectures used in the Experiments}



\textbf{Data pre-processing and post-processing:} For the experiments on Cifar-10 and Cifar-100 datasets, we used two standard data augmentation techniques which are horizontal flipping and translation by 4 pixels \cite{res_net,SN}. 

For the experiments on Imagenet dataset, we followed the data augmentation methods suggested in \cite{res_net}. In addition, we used both the scale and aspect ratio augmentation used in \cite{go_deeper1}. For color augmentation, we used the photometric distortions \cite{Howard13} and standard color augmentation \cite{res_net}. Moreover, we used random sampling of  $224 \times 224$ crops or their horizontal flips with the normalized data obtained by subtracting per-pixel mean. In the bottleneck blocks, stride 2 is used for the $A_l=B_l=3$ weights. Moreover, Euclidean gradient decays are employed for all the weights.

\textbf{Acceleration methods:} In this section, we employed state-of-the-art acceleration methods \cite{on_mom} modularly in Algorithm~1 for implementation of the CNNs as suggested in the reference works \cite{res_net,SN,ooAAAI18}. In this work, we consider employment of acceleration methods on the ambient Euclidean space and collections of POMs as suggested in \cite{ooAAAI18}. For this purpose, momentum and Euclidean gradient decay methods are employed on the Euclidean gradient $ \gr_E \; \mathcal{L}(\omega_{g,l}^{t})$ using $\mu_t :=  q \Big (  \gr_E \; \mathcal{L}(\omega_{g,l}^{t}),\mu_t,\Theta \Big)$. We can employ state-of-the-art acceleration methods \cite{on_mom} modularly in this step. Thus, momentum was employed with the Euclidean gradient decay using
\vspace{-0.05cm}
\begin{equation}
q \Big (  \gr_E \; \mathcal{L}(\omega_{g,l}^{t}),\mu_t,\Theta \Big) = \theta_{\mu} \mu_t - \theta_E  \gr_E \; \mathcal{L}(\omega_{g,l}^{t}),
\label{mom_decay}	
\vspace{-0.1cm}
\end{equation} 
where $\theta_{\mu} \in \Theta$ is the parameter employed on the momentum variable $\mu_t$. We consider $\theta_E \in \Theta$ as the decay parameter for the Euclidean gradient. In the experiments, we used $\theta_{\mu} = \theta_E = 0.9$.

\textbf{Architectural Details of CNNs:} In the experiments, we used the same hyper-parameters of CNN architectures (e.g. number of channels, layers, weight sizes, stride and padding parameters) and their implementation provided by the authors of the compared works for training of CNNs using our proposed SGD method, for a fair comparison with base-line methods. Differences between the implementations and hyper-parameters are explained below. In other words, we just implemented the SGD algorithm of the provided CNN implementations using our proposed SGD method. More precisely, we used the following implementations for comparison:

\begin{itemize}[leftmargin=*]		
	\item RCD and RSD: We used the Residual networks with constant and stochastic depth using the same configuration hyper-parameters (see below for number of weights used in the architectures) and code given in \cite{SN}.
	
	\item Residual Networks (Resnets): We re-implemented residual networks with the same configuration and training hyper-parameters (see below for number of weights used in the architectures) given in \cite{res_net,ooAAAI18}.
	
	\item Squeeze-and-Excitation networks implemented for Resnets with 50 layers (SENet-Resnet-50): We re-implemented residual networks with the same configuration and training hyper-parameters (see below for number of weights used in the architectures) given in \cite{senet}.
	
\end{itemize}

In order to construct collections of weights belonging to four spaces (Euc., Sp, St and Ob) using WSS, we increase the number of weights used in CNNs to 24 and its multiples as follows;


$\bullet$ Resnet with 18 Layers (Table~6 in this text): 72 filters at the first and second, 144 filters at the third, 288 filters at the fourth, and 576 filters at the fifth convolution blocks \cite{res_net}.

$\bullet$ Resnet with 44 Layers (Table~7 in this text): 24 filters for 15 layers, 48 filters for 14 layers, 96 filters for 14  \cite{res_net}.

$\bullet$ Resnets with constant depth (RCD) and stochastic depth (RSD) with 110 layers (Table~2 in the main text and Table~8 in this text): 24, 48 and 72 filters at the first, second, and the third convolution blocks \cite{SN}.

$\bullet$ Resnet-50 and SENet-Resnet-50 (Table~1 in the main text): Configurations of Resnet-50 and SENet-Resnet-50 are given in Table~\ref{res50} and Table~\ref{seres50}, respectively.

\begin{table*}[t]
	\caption{Configuration details of the Resnet-50 used for the experiments given in Table~1 in the main text.}
	\begin{center}
		\begin{tabular}{|c|c|}
			\hline
			\multicolumn{1}{|l|}{Output Size} & Resnet-50 \\ \hline
			\multicolumn{ 1}{|c|}{$112\times 112$} & Kernel size: $7\times7$,  Number of convolution weights: 64, Stride 2 \\   \hline
			\multicolumn{ 1}{|c|}{$56 \times 56$} & $3 \times 3$ Max Pooling,  Stride 2 \\  
			\multicolumn{ 1}{|c|}{} & 3 Residual Blocks with the Following Convolution Kernels: \\  
			\multicolumn{ 1}{|c|}{} & 72 convolution weights of size $1 \times 1$ \\  
			\multicolumn{ 1}{|c|}{} & 72 convolution weights of size $3 \times 3$ \\  
			\multicolumn{ 1}{|c|}{} & 264 convolution weights of size $1 \times 1$ \\   \hline
			\multicolumn{ 1}{|c|}{$28 \times 28$} & \\  
			\multicolumn{ 1}{|c|}{} & 4 Residual Blocks with the Following Convolution Kernels: \\  
			\multicolumn{ 1}{|c|}{} & 144 convolution weights of size $1 \times 1$ \\  
			\multicolumn{ 1}{|c|}{} & 144 convolution weights of size $3 \times 3$ \\  
			\multicolumn{ 1}{|c|}{} & 528 convolution weights of size $1 \times 1$ \\   \hline
			\multicolumn{ 1}{|c|}{$14 \times 14$} &  \\  
			\multicolumn{ 1}{|c|}{} & 6 Residual Blocks with the Following Convolution Kernels: \\  
			\multicolumn{ 1}{|c|}{} & 264 convolution weights of size $1 \times 1$ \\  
			\multicolumn{ 1}{|c|}{} & 264 convolution weights of size $3 \times 3$ \\  
			\multicolumn{ 1}{|c|}{} & 1032 convolution weights of size $1 \times 1$ \\   \hline
			\multicolumn{ 1}{|c|}{$7 \times 7$} &  \\  
			\multicolumn{ 1}{|c|}{} & 3 Residual Blocks with the Following Convolution Kernels: \\  
			\multicolumn{ 1}{|c|}{} & 528 convolution weights of size $1 \times 1$ \\  
			\multicolumn{ 1}{|c|}{} & 528 convolution weights of size $3 \times 3$ \\  
			\multicolumn{ 1}{|c|}{} & 2064 convolution weights of size $1 \times 1$ \\   \hline
			\multicolumn{ 1}{|c|}{$1\times1$} & Global Average Pooling \\  
			\multicolumn{ 1}{|c|}{} & Fully connected layer\\ 
			\multicolumn{ 1}{|c|}{} & Softmax \\ \hline
		\end{tabular}
	\end{center}
	\label{res50}
\end{table*}

\begin{table*}[t]
	\caption{Configuration details of the SENet-Resnet-50 used for the experiments given in Table~1 in the main text.}
	\begin{center}
		\begin{tabular}{|c|c|}
			\hline
			\multicolumn{1}{|l|}{Output Size} & Resnet-50 \\ \hline
			\multicolumn{ 1}{|c|}{$112\times 112$} & Kernel size: $7\times7$,  Number of convolution weights: 64, Stride 2 \\   \hline
			\multicolumn{ 1}{|c|}{$56 \times 56$} & $3 \times 3$ Max Pooling,  Stride 2 \\  
			\multicolumn{ 1}{|c|}{} & 3 Residual Blocks with the Following Convolution Kernels: \\  
			\multicolumn{ 1}{|c|}{} & 72 convolution weights of size $1 \times 1$ \\  
			\multicolumn{ 1}{|c|}{} & 72 convolution weights of size $3 \times 3$ \\  
			\multicolumn{ 1}{|c|}{} & 264 convolution weights of size $1 \times 1$ \\   
			\multicolumn{ 1}{|c|}{} & Fully connected layer with weights of size $24 \times 264$ \\ \hline
			\multicolumn{ 1}{|c|}{$28 \times 28$} & \\  
			\multicolumn{ 1}{|c|}{} & 4 Residual Blocks with the Following Convolution Kernels: \\  
			\multicolumn{ 1}{|c|}{} & 144 convolution weights of size $1 \times 1$ \\  
			\multicolumn{ 1}{|c|}{} & 144 convolution weights of size $3 \times 3$ \\  
			\multicolumn{ 1}{|c|}{} & 528 convolution weights of size $1 \times 1$ \\ 
			\multicolumn{ 1}{|c|}{} & Fully connected layer with weights of size $48 \times 528$ \\ \hline
			\multicolumn{ 1}{|c|}{$14 \times 14$} &  \\  
			\multicolumn{ 1}{|c|}{} & 6 Residual Blocks with the Following Convolution Kernels: \\  
			\multicolumn{ 1}{|c|}{} & 264 convolution weights of size $1 \times 1$ \\  
			\multicolumn{ 1}{|c|}{} & 264 convolution weights of size $3 \times 3$ \\  
			\multicolumn{ 1}{|c|}{} & 1032 convolution weights of size $1 \times 1$ \\   
			\multicolumn{ 1}{|c|}{} & Fully connected layer with weights of size $72 \times 1032$ \\ \hline
			\multicolumn{ 1}{|c|}{$7 \times 7$} &  \\  
			\multicolumn{ 1}{|c|}{} & 3 Residual Blocks with the Following Convolution Kernels: \\  
			\multicolumn{ 1}{|c|}{} & 528 convolution weights of size $1 \times 1$ \\  
			\multicolumn{ 1}{|c|}{} & 528 convolution weights of size $3 \times 3$ \\  
			\multicolumn{ 1}{|c|}{} & 2064 convolution weights of size $1 \times 1$ \\  
			\multicolumn{ 1}{|c|}{} & Fully connected layer with weights of size $144 \times 2064$ \\ \hline
			\multicolumn{ 1}{|c|}{$1\times1$} & Global Average Pooling \\  
			\multicolumn{ 1}{|c|}{} & Fully connected layer\\ 
			\multicolumn{ 1}{|c|}{} & Softmax \\ \hline
		\end{tabular}
	\end{center}
	\label{seres50}
\end{table*}


\textbf{Scaling of weights:} 
We use $\Re_{l}^t$ for scaling of weights and identification of component weight manifolds of POMs. As we mentioned in the main text, for instance, $\Re_{l}^t$ is computed and used as the radius of the sphere. More precisely, we initialize weights $\omega \in \mathcal{M}_{\iota}$ that belong to the sphere $\mathcal{M}_{\iota} \equiv \mathcal{S}(A_l,B_l)$ subject to the constraint $\| \omega \|^2_{F} = \Re_{l}^t$ by constructing a scaled sphere 
\begin{equation}
\mathbb{S}^{A_l B_l-1} \triangleq \mathcal{S}_{\Re_{l}^t}(A_l,B_l) =  \{\omega \in \mathbb{R}^{A_l \times B_l} : \| \omega \|^2_{F} = \Re_{l}^t \}.
\end{equation}
The other manifolds (the oblique and the Stiefel manifolds) are identified, and the weights that belong to the manifolds are initialized, appropriately, following the aforementioned methods. Then, projection of gradients, exponential maps and retractions which are determined according to manifold structure of weight spaces (see Table~\ref{tab:tangent_spaces} and Table~\ref{tab:groups_manifolds}), are updated accordingly by $\Re_{l}^t$. For example, for the scaled sphere $\mathcal{S}_{\Gamma_l^t}(A_l,B_l)$, we compute the projection of gradients by ${(I\Re_{l}^t-\omega \omega^T)\mu}$, and the exponential map by 
\begin{equation}
\exp_{\omega}(v) ={\omega}\cos(\|v \|_F \Re_{l}^t) +  \Re_{l}^t \frac{v}{\| v \|_F}\sin(\| v \|_F \Re_{l}^t) .
\end{equation}

\newpage

\subsection{Employment of Weight Set Splitting Scheme (WSS) in the Experiments:}



\begin{figure*}[t!]
	\centering
	\includegraphics[scale=0.450]{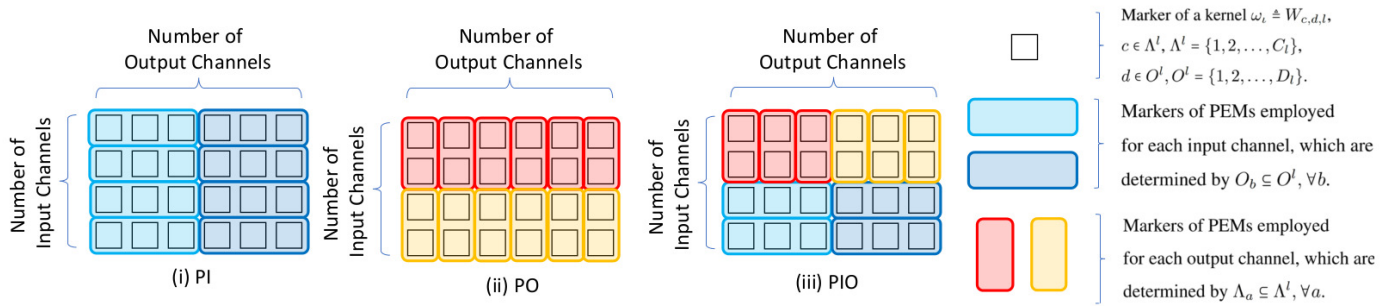}%
	\caption{An illustration for employment of the proposed PI, PO and PIO strategies at the $l^{th}$ layer of a CNN.}
	\label{fig_block1}
\end{figure*}

Recall that, at each $l^{th}$ layer, we compute a weight ${\omega_{\iota} \triangleq W_{c,d,l}}$, ${c \in \Lambda^l}$,  $\Lambda^l=\{1,2,\ldots,C_l\}$, ${d \in O^l}$, ${O^l=\{1,2,\ldots,D_l \}}$. We first choose $\mathfrak{A}$ subsets of indices of input channels ${\Lambda_a \subseteq \Lambda^l}, a=1,2,\ldots,\mathfrak{A}$, and $\mathfrak{B}$ subsets of indices of output channels $O_b \subseteq O^l, b=1,2,\ldots,\mathfrak{B}$, such that $\Lambda^l = \bigcup \limits _{a=1} ^\mathfrak{A} \Lambda_a$ and $O^l = \bigcup \limits _{b=1} ^\mathfrak{B} O_b$. We determine indices of weights belonging to different groups using the following three schemes:

\begin{enumerate}[leftmargin=*] 
	\item POMs for input channels (PI): For each $c^{th}$ input channel, we construct $\mathcal{I}_{\mathcal{G}_l} = \bigcup \limits _{c=1} ^{C_l} \mathcal{I}_{{G}_l} ^c   $, where  ${\mathcal{I}_{{G}_l} ^c =  O_b \times \{c\} }$ and the Cartesian product ${O_b \times \{c\}} $ preserves the input channel index, $\forall b,c$ (see Figure~\ref{fig_block1}). 	
	\item POMs for output channels (PO): For each $d^{th}$ output channel, we construct $\mathcal{I}_{\mathcal{G}_l} = \bigcup \limits _{d=1} ^{D_l} \mathcal{I}_{{G}_l} ^d  $, where   ${\mathcal{I}_{{G}_l} ^d = \Lambda_a \times \{d\}  }$ and the Cartesian product $\Lambda_a \times \{d\} $ preserves the output channel index, $\forall a,d$ (see Figure~\ref{fig_block1}). 	
	\item POMs for input and output channels (PIO): In PIO, we construct $\mathcal{I}_{l}^{a,b} =  \mathcal{I}_{l}^{a} \cup  \mathcal{I}_{l}^{b}$, where $ \mathcal{I}_{l}^{a} = \{ \Lambda_a \times a \}$, and $ {\mathcal{I}_{l}^{b} = \{ O_b \times b\} }$  such that $\mathcal{I}_{\mathcal{G}_l} = \bigcup \limits _{a=1, b=1} ^{\mathfrak{A},\mathfrak{B}} \mathcal{I}_{l} ^{a,b}$ (see Figure~\ref{fig_block1}).	
\end{enumerate} 

\textbf{Illustrative Examples of Employment of PI, PO and PIO}

A comparative and illustrative example for comparison of PI, PO and PIO is given in Figure~\ref{fig_block1}.

\begin{example}
	Suppose that we have a weight tensor of size $3 \times 3 \times 4 \times 6$ where the number of input and output channels is $4$ and $6$. In total, we have ${4*6=24}$ weight matrices of size $3 \times 3$. An example of construction of an collection of POMs is as follows. 
	\begin{enumerate}[leftmargin=*] 
		\item PIO: We split the set of 24 weights into 10 subsets. For 6 output channels, we split the set of weights corresponding to 4 input channels into 3 subsets. We choose the sphere (Sp) for {\color{blue} 2 subsets} each containing 3 weights (depicted by {\color{blue} light blue rectangles}), and {\color{red} 3 subsets} each containing 2 weights (depicted by {\color{red} red rectangles}). We choose the Stiefel manifold (St) similarly for the remaining subsets. Then, our ensemble contains 5 POMs of St and 5 POMs of Sp.
		\item PI: For each of 4 input channels, we split a set of 6 weights associated with 6 output channels into two subsets of 3 weights. Choosing the sphere (Sp) for the first subset, we construct a POM  as a product of 3 Sp. That is, each of 3 component manifolds ${\mathcal{M}_{\iota}}, {\iota  = 1,2,3}$, of the POM is a sphere. Similarly, choosing the Stiefel (St) for the second subset, we construct another POM as a product of 3 St (each of 3 component manifolds ${\mathcal{M}_{\iota}}, \iota  = 1,2,3$, of the second POM is a Stiefel manifold.). Thus, at this layer, we construct an collection of 4 POMs of 3 St and 4 POMs of 3 Sp.
		\item PO: For each of 6 output channels, we split a set of 4 weights corresponding to the input channels into two subsets of 2 weights. We choose the Sp for the first subset, and we construct a POM as a product of 2 Sp using. We choose the St for the second subset, and we construct a POM as a product of 2 St. Thereby, we have an collection consisting of 6 POMs of St and 6 POMs of Sp.
		
	\end{enumerate}
\end{example}

\begin{figure*}[ht]
	\centering
	\includegraphics[width=5.70in]{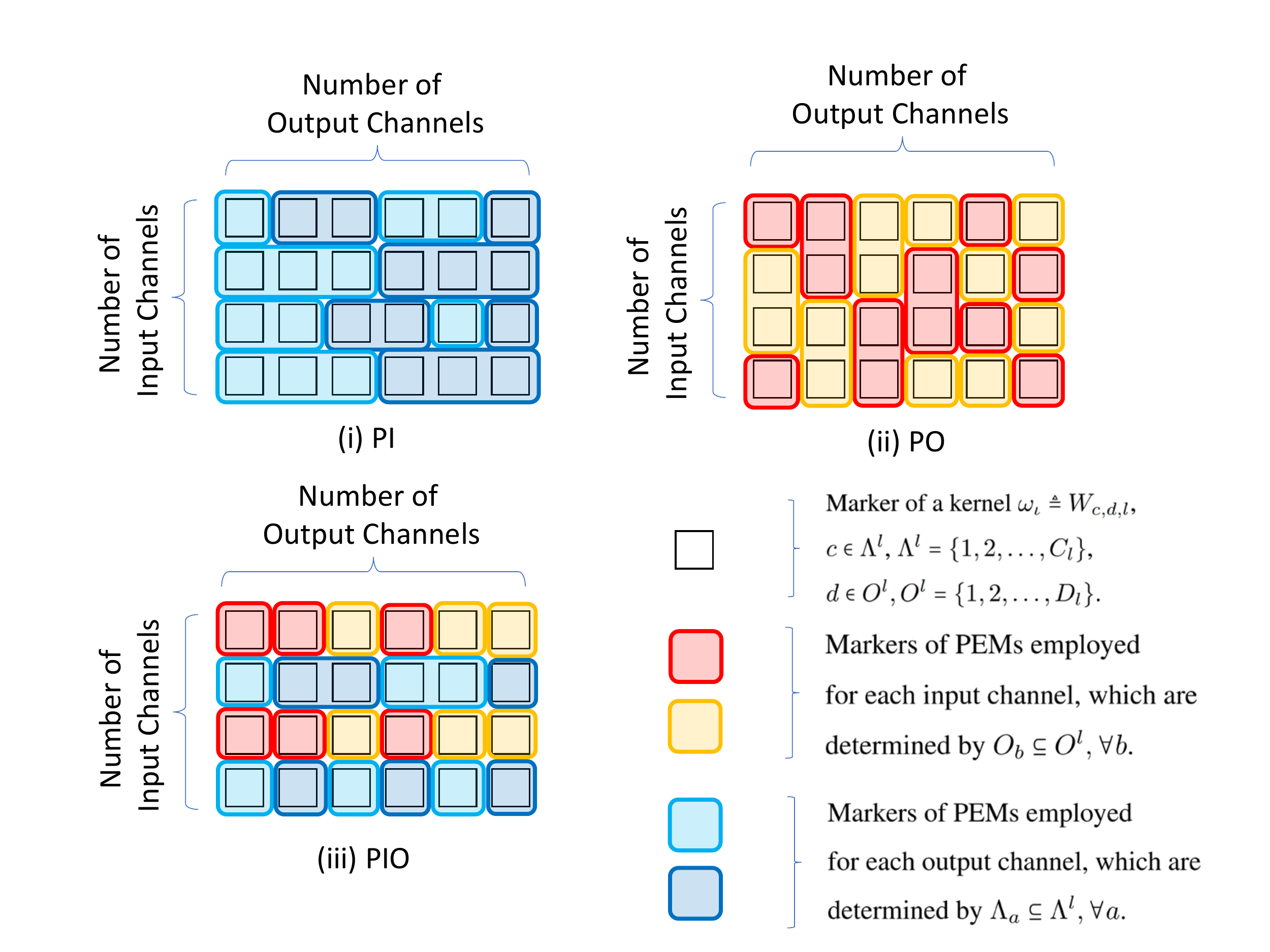}%
	\caption{An illustration of employment of the proposed PI, PO and PIO collection strategies at the $l^{th}$ layer of a CNN. In Section~4.2, we randomly selected indices of weights, i.e. subsets of input and output channels, according to the uniform distribution. In this example, we suppose that there are four input and six output channels. Then, 24 convolution weights are computed on in two different POMs.}
	\label{fig_random}
\end{figure*}

In the experiments, indices of weights for PI, PO and PIO are randomly selected. An illustration of the selection method is given in Figure~\ref{fig_random}.

\newpage 

\textbf{Notation used in the Tables}

\begin{enumerate}
	\item Sp/Ob/St: Kernels employed on each input and output channel are defined to reside on the sphere, oblique and Stiefel manifold, respectively.
	
	\item POMs of Sp/Ob/St: Kernels employed on all input and output channels are defined to reside on a POM of Sp/Ob/St.
	
	\item PI/PO/PIO for POMs of Sp/Ob/St: Ensembles of POMs of Sp/Ob/St are computed using the schemes PI/PO/PIO.

	\item Results for Manifold$_1$ + Manifold$_2$: Results are computed for collections of POMs of Manifold$_1$ and Manifold$_2$.
	
	\item Results for Manifold$_1$ + Manifold$_2$ + Manifold$_3$: Results are computed for collections of POMs of Manifold$_1$, Manifold$_2$ and Manifold$_3$.
	
	\item Results for  Manifold$_1$ + Manifold$_2$ + Manifold$_3$ + Manifold$_4$: Results are computed for collections of POMs of Manifold$_1$, Manifold$_2$, Manifold$_3$ and Manifold$_4$.
	
\end{enumerate}

\section{Additional Results}

\subsection{Analyses using Resnets with Different Number of Layers}

In this subsection, we give additional results for image classification using Cifar-10 and Imagenet datasets for different networks such as Resnets with 18 and 44 layers (Resnet-18 and Resnet-44), 110-layer Resnets with constant depth (RCD) and stochastic depth (RSD)  with data augmentation (DA) and without using data augmentation (w/o DA).

\begin{table}[t]
	\centering
	\vspace{-0.28cm}			
	\caption{Results for Resnet-18 which are trained using the Imagenet for single crop validation error rate (\%).}		
	\begin{tabular}{C{4.85cm} C{2.70cm}}
		\toprule
		\toprule
		
		\textbf{Model} & \textbf{Top-1 Error (\%)} \\
		
		\midrule
		\midrule
		Euc. \cite{ooAAAI18} & 30.59\\
		Euc. $\dagger$ & {\color{red} 30.31}\\
		Sp/Ob/St\cite{ooAAAI18}  & 29.13/28.97/{{28.14}}\\
		Sp/Ob/St $\dagger$  & 28.71/28.83/{{28.02}}\\	
		POMs of Sp/Ob/St & 28.70/28.77/{{28.00}}\\	
		PI for POMs of Sp/Ob/St & 28.69/28.75/{{27.91}}\\				
		PI (Euc.+Sp/Euc.+St/Euc.+Ob) & 30.05/29.81/29.88  \\		
		PI (Sp+Ob/Sp+St/Ob+St) & 28.61/28.64/28.49  \\
		PI (Sp+Ob+St/Sp+Ob+St+Euc.)  & 27.63/27.45  \\	
		PO for POMs of Sp/Ob/St & 28.67/28.81/{{27.86}}\\						
		PO (Euc.+Sp/Euc.+St/Euc.+Ob) & 29.58/29.51/29.90  \\		
		PO (Sp+Ob/Sp+St/Ob+St) & 28.23/28.01/28.17  \\
		PO (Sp+Ob+St/Sp+Ob+St+Euc.)  & 27.81/27.51  \\
		PIO for POMs of Sp/Ob/St & 28.64/28.72/{{27.83}}\\								
		PIO (Euc.+Sp/Euc.+St/Euc.+Ob) & 29.19/28.25/28.53  \\		
		PIO (Sp+Ob/Sp+St/Ob+St) & 28.14/27.66/27.90  \\
		PIO (Sp+Ob+St/Sp+Ob+St+Euc.)  & 27.11/{\color{blue} 27.07}  \\			
		\bottomrule
		\bottomrule
	\end{tabular}%
	\label{tab:imagenet}%
\end{table}%

We give classification performance of Resnets with 18 layers (Resnet-18) employed on the Imagenet  in Table~\ref{tab:imagenet}. The results show that performance of CNNs are boosted by employing collections of POMs (denoted by PIO for POMs) using FG-SGD compared to the employment of baseline Euc. We observe that POMs of component manifolds of identical geometry (denoted by POMs of Sp/St/Ob), and their collections (denoted by PIO for POMs of Sp/St/Ob)  provide better performance compared to employment of individual component manifolds (denoted by Sp/Ob/St) \cite{ooAAAI18}. For instance, we obtain $28.64\%$, $28.72\%$ and $27.83\%$ error using PIO for POMs of Sp, Ob and St in Table~\ref{tab:imagenet}, respectively. However, the error obtained using Sp, Ob and St is $28.71\%$, $28.83\%$ and $28.02\%$, respectively. We observe {\color{blue} $3.24\%$} boost by construction of an collection of four manifolds (Sp+Ob+St+Euc.) using the  PIO scheme in Table~\ref{tab:imagenet} ({\color{blue} $27.07\%$}). In other words, collection methods boost the performance of large-scale CNNs more for large-scale datasets (e.g. Imagenet) consisting of larger number of samples and classes compared to the performance of smaller CNNs employed on smaller datasets (e.g. Cifar-10). This result can be attributed to enhancement of sets of features learned using multiple constraints.

In addition, we obtain $0.28\%$ and $2.06\%$ boost of the performance by collection of the St with Euc. ($6.77\%$ and $28.25\%$ using PIO for Euc.+St, respectively) for the experiments on the Cifar-10 and Imagenet datasets using the  PIO scheme in Table~\ref{tab:res10} and Table~\ref{tab:imagenet}, respectively. Moreover, we observe that construction of collections using Ob performs better for PI compared to PO. For instance, we observe that PI for POMs of Ob provides $6.81\%$ and $28.75\%$ while PO for POMs of Ob provides $6.83\%$ and $28.81\%$ in Table~\ref{tab:res10} and Table~\ref{tab:imagenet}, respectively.   We may associate this result with the observation that weights belonging to Ob are used for feature selection and modeling of texture patterns with high performance \cite{oblq,oo16}. However, collections of St and Sp perform better for PO ($6.59\%$ and $28.01\%$ in Table~\ref{tab:res10} and Table~\ref{tab:imagenet}) compared to PI ($6.67\%$ and $28.64\%$ in Table~\ref{tab:res10} and Table~\ref{tab:imagenet}) on weights employed on output channels. 

It is also observed that  PIO performs better than PI and PO in all the experiments. We observe {\color{blue} $3.24\%$} boost by construction of an collection of four manifolds (Sp+Ob+St+Euc.) using the  PIO scheme in Table~\ref{tab:imagenet} ({\color{blue} $27.07\%$}). In other words, collection methods boost the performance of large-scale CNNs more for large-scale datasets (e.g. Imagenet) consisting of larger number of samples and classes compared to the performance of smaller CNNs employed on smaller datasets (e.g. Cifar-10). This result can be attributed to enhancement of sets of features learned using multiple constraints.

\begin{table}[t]
	\centering
	\caption{Results for Resnet-44 on the Cifar-10 with DA.}
	\begin{tabular}{C{4.95cm}C{2.5cm}}
		\toprule
		\toprule
		
		\textbf{Model} & \textbf{Class. Error(\%)} \\
		\midrule
		\midrule
		
		Euc. \cite{res_net}  & 7.17  \\
		Euc. \cite{ooAAAI18} & 7.16  \\
		Euc. $\dagger$ & {\color{red} 7.05}  \\
		Sp/Ob/St \cite{ooAAAI18} & 6.99/6.89/{{6.81}}\\
		Sp/Ob/St  $\dagger$ & 6.84/6.87/{ {6.73}}\\
		POMs of	Sp/Ob/St   & 6.81/6.85/{ {6.70}}\\
		PI for POMs of	Sp/Ob/St   & 6.82/6.81/{ {6.70}}\\
		PI (Euc.+Sp/Euc.+St/Euc.+Ob) & 6.89/6.84/6.88  \\
		PI (Sp+Ob/Sp+St/Ob+St) & 6.75/6.67/6.59 \\	
		PI (Sp+Ob+St/Sp+Ob+St+Euc.)  & 6.31/6.34  \\	
		PO for POMs of	Sp/Ob/St  & 6.77/6.83/{ {6.65}}\\	
		PO (Euc.+Sp/Euc.+St/Euc.+Ob) & 6.85/6.78/6.90  \\		
		PO (Sp+Ob/Sp+St/Ob+St) & 6.62/6.59/6.51  \\	
		PO (Sp+Ob+St/Sp+Ob+St+Euc.)  & 6.35/6.22 \\		
		PIO for POMs of	Sp/Ob/St   & 6.71/6.73/{ {6.61}}\\					
		PIO (Euc.+Sp/Euc.+St/Euc.+Ob) & 6.95/6.77/6.82  \\		
		PIO (Sp+Ob/Sp+St/Ob+St) & 6.21/6.19/6.25  \\
		PIO (Sp+Ob+St/Sp+Ob+St+Euc.)  & 5.95/{\color{blue} 5.92 } \\		
		\bottomrule
		\bottomrule
	\end{tabular}%
	\label{tab:res10}%
\end{table}%

\begin{table*}[ht]
	\centering
	\caption{Classification error (\%) for training 110-layer Resnets with constant depth (RCD) and Resnets with stochastic depth (RSD) using the PIO scheme on  Cifar-100, with data augmentation (w. DA) and without using DA (w/o DA).}	
	\begin{tabular}{C{4.8cm}|C{2.750cm}|C{2.69cm}|}
		\toprule
		\toprule
		\multicolumn{1}{c}{\textbf{Model}} & \multicolumn{1}{c}{\textbf{Cifar-100 w. DA}} & \multicolumn{1}{c}{\textbf{Cifar-100 w/o DA}} \\
		\bottomrule
		RCD \cite{DCCN}  & 27.22   & 44.74 \\
		(Euc.) $\dagger$  &{\color{red} 27.01}  & {\color{red} 44.65 }\\		
		Sp/Ob/St (\cite{ooAAAI18}) & 26.44/25.99/{{25.41}} & 42.51/42.30/{{40.11}} \\
		Sp/Ob/St $\dagger$   & 26.19/25.87/{{25.39}} & 42.13/42.00/{{39.94}} \\
		POMs of Sp/Ob/St  & 25.93/25.74/{{25.18}} & 42.02/42.88/{{39.90}} \\		
		PIO (Euc.+Sp/Euc.+St/Euc.+Ob)    & 25.57/25.49/25.64  & 41.90/41.37/41.85 \\	
		PIO   (Sp+Ob/Sp+St/Ob+St) & 24.71/24.96/24.76  & 41.49/40.53/40.34  \\		
		PIO (Sp+Ob+St/Sp+Ob+St+Euc.)  & 23.96/{\color{blue} 23.79 }  & 39.53/ {\color{blue} 39.35 } \\
		\bottomrule
		RSD \cite{DCCN}  & 24.58  & 37.80  \\
		Euc. $\dagger$   & {\color{red} 24.39} & {\color{red} 37.55 } \\		
		Sp/Ob/St \cite{ooAAAI18}  & 23.77/23.81/{{23.16}} & 36.90/36.47/{{35.92}} \\
		Sp/Ob/St $\dagger$  & 23.69/23.75/{{23.09}} & 36.71/36.38/{{35.85}} \\
		POMs of Sp/Ob/St  & 23.51/23.60/{{23.85}} & 36.40/36.11/{{35.53}} \\
		PIO   (Euc.+Sp/Euc.+St/Euc.+Ob)   & 23.69/23.25/23.32  & 35.76/35.55/35.81 \\	
		PIO   (Sp+Ob/Sp+St/Ob+St)  & 22.84/22.91/22.80  & 35.66/35.01/35.35 \\		
		PIO (Sp+Ob+St/Sp+Ob+St+Euc.)  & 22.19/{\color{blue} 22.03}  & 34.49/{\color{blue} 34.25} \\
		\bottomrule
		\bottomrule
		
	\end{tabular}%
	\label{tab:rcd100}%
\end{table*}%

\begin{table*}[ht]
	\centering
	\caption{Classification error (\%) for training 110-layer Resnets with constant depth (RCD) and Resnets with stochastic depth (RSD) using the PIO scheme on the Cifar-10, with and without using DA.}	
	\begin{tabular}{C{4.8cm}|C{2.7cm}|C{2.69cm}|}
		\toprule
		\toprule
		\multicolumn{1}{c}{\textbf{Model}} & \multicolumn{1}{c}{\textbf{Cifar-10 w. DA}} & \multicolumn{1}{c}{\textbf{Cifar-10 w/o DA}}  \\
		\bottomrule
		RCD \cite{DCCN}  & 6.41 &  13.63 \\
		(Euc.) $\dagger$  & {\color{red} 6.30}  &{\color{red} 13.57}  \\		
		Sp/Ob/St (\cite{ooAAAI18}) & 6.22/6.07/{{5.93}} & 13.11/12.94/{{12.88}}  \\
		Sp/Ob/St $\dagger$  & 6.05/6.03/{{5.91}} & 12.96/12.85/{{12.79}} \\
		POMs of Sp/Ob/St & 6.00/6.01/{{5.86}}  & 12.74/12.77/{{12.74}}  \\		
		PIO for POMs of Sp/Ob/St  & 5.95/5.91/{{5.83}}  & 12.71/12.72/{{12.69}}  \\
		PIO (Euc.+Sp/Euc.+St/Euc.+Ob)  & 6.03/5.99/6.01   & 12.77/12.21/12.92 \\	
		PIO   (Sp+Ob/Sp+St/Ob+St)  & 5.97/5.86/5.46   & 11.47/11.65/ 11.51   \\		
		PIO (Sp+Ob+St/Sp+Ob+St+Euc.)  & 5.25/{\color{blue} 5.17}   & 11.29/{\color{blue} 11.15}  \\
		\bottomrule
		RSD \cite{DCCN}   & 5.23  & 11.66   \\
		Euc. $\dagger$   & {\color{red} 5.17 }& {\color{red} 11.40} \\		
		Sp/Ob/St \cite{ooAAAI18} & 5.20/5.14/{{4.79}} & 10.91/10.93/{{10.46}} \\
		Sp/Ob/St $\dagger$ & 5.08/5.11/{{4.73}}  & 10.52/10.66/{{10.33}} \\
		POMs of Sp/Ob/St & 5.05/5.08/{{4.69}}   & 10.41/10.54/{{10.25}}  \\
		PIO for POMs of Sp/Ob/St & 4.95/5.03/{{4.62}}   & 10.37/10.51/{{10.19}}  \\		
		PIO   (Euc.+Sp/Euc.+St/Euc.+Ob)  & 5.00/5.08/5.14   & 10.74/10.25/10.93   \\	
		PIO   (Sp+Ob/Sp+St/Ob+St)  & 4.70/4.58/4.90   & 10.13/10.24/10.06   \\		
		PIO (Sp+Ob+St/Sp+Ob+St+Euc.)  & {\color{blue} 4.29}/4.31   &  {\color{blue} 9.52}/9.56  \\
		\bottomrule
		\bottomrule
		
	\end{tabular}%
	\label{tab:rcd}%
\end{table*}%

\begin{table*}[t]
	\centering
	\caption{Mean $\pm$ standard deviation of classification error (\%) are given for results obtained using SENet-Resnet-101, and 110-layer Resnets with constant depth (RCD) on Cifar-100.}		
	\begin{tabular}{C{9.5cm}|C{2.7cm}|}
		\toprule
		\toprule		
		\textbf{Model} \textbf{Cifar-100 with DA (110 layer RCD)} & \textbf{Error}  \\		
		Euc. $\dagger$ & {\color{red}  27.01 $\pm$ 0.47}  \\
		St & 25.39  $\pm$ 0.40 \\	
		POMs of St  & 25.18  $\pm$ 0.34\\		
		PIO (Sp+Ob+St) & 23.96  $\pm$ 0.28 \\			
		PIO (Sp+Ob+St+Euc.) & {\color{blue} 23.79  $\pm$ 0.15} \\			
		\bottomrule
		(Additional results) \textbf{Cifar-100 with DA (SENet-Resnet-101)} & \textbf{Error} \\
		Euc. $\dagger$  &  {\color{red} 19.93 $\pm$ 0.51} \\		
		PIO (Sp+Ob+St) &  18.96 $\pm$  0.27		\\
		PIO (Sp+Ob+St+Euc.)  &   {\color{blue} 18.54 $\pm$ 0.16}		\\
		\bottomrule		
		\bottomrule
	\end{tabular}%
	\label{tab:summary}%
\end{table*}%

In Table~\ref{tab:rcd100}, we analyze the performance of larger CNNs consisting of 110 layers on Cifar-100 with and without using DA. We implemented the experiments 10 times and provided the average performance. 
We observe that sets boost the performance of CNNs that use DA methods more compared to the performance of CNNs without using DA. For instance, PIO of all manifolds ({\color{blue} 39.35$\%$}) outperform baseline ({\color{red} 44.65$\%$}) by $5.3\%$ without using DA, while those ({\color{blue} 23.79$\%$}) obtained using DA outperform baseline ({\color{red} 27.01$\%$}) by $3.22\%$ for RCD. Additional results for different CNNs using Imagenet and Cifar-10, and a comparison with vanilla network sets are given in this supplemental material.  

\newpage

\subsection{Comparison with Vanilla Network Ensembles}

Our method fundamentally differs from network ensembles. In order to analyze the results for network ensembles of CNNs, we employed an ensemble method \cite{res_net} by \textit{voting of decisions} of Resnet 44 on Cifar 10. When CNNs trained on individual Euc, Sp, Ob, and St are ensembled using voting, we obtained $7.02\%$  (Euc+Sp+Ob+St) and $6.85\%$ (Sp+Ob+St) errors (see Table 1 for comparison). In our analyses of ensembles (PI, PO and PIO), each POM contains $\frac{N_l}{M}$ weights, where $N_l$ is the number of weights used at the $l^{th}$ layer, and $M$ is the number of POMs. When each CNN in the ensemble was trained using an individual manifold which contains $\frac{1}{4}$ of weights (using $M=4$ as utilized in our experiments), then we obtained $11.02\%$ (Euc), $7.76\%$ (Sp), $7.30\%$ (Ob), $7.18\%$ (St), $9.44\%$ (Euc+Sp+Ob+St) and $7.05\%$ (Sp+Ob+St) errors. Thus, our proposed methods outperform ensembles constructed by voting.

\subsection{Analyses for Larger DNNs with Large Scale Image Datasets}


We give the results for Cifar-100 obtained using data augmentation denoted by with DA in Table~\ref{tab:summary}.Cifar-100 dataset consist of $5\times 10^4$  training and $10^4$ test images belonging to 100 classes. 



In Table~\ref{tab:summary}, we provide results using the state-of-the-art Squeeze-and-Excitation (SE) blocks \cite{senet} implemented for Resnets with 110 layers (Resnet-110) on Cifar-100. We run the experiments 3 times and provide the average performance.

In the second set of experiments, we perform separable convolution operations using the proposed weight splitting scheme.  We compare the results using various popular separable convolution schemes, such as depth-wise and channel-wise convolution implemented using state-of-the-art DNNs such as ResNext with 50 layers (ResNext-50) \cite{resnext}, MobileNet v2 with 21 layers (Mobilenet) \cite{Sandler} and 50 layer Resnets with hierarchical filtering using 4 roots (DeepRoots) \cite{Ioanno}. The results obtained using PIO with (Sp+Ob+St+Euc.) with the separable convolution scheme proposed in the corresponding related work are denoted by PIO-SOSE. The results obtaied using PIO with (Sp+Ob+St+Euc.) with our proposed WSS are denoted by PIO-SOSE-WSS.

\begin{table}[t]
	\centering
	\caption{Analysis of classification error (\%) of state-of-the-art DNNs which employ separable convolutions on Imagenet  dataset.}		
	\begin{tabular}{c c}
		\toprule
		\toprule		
		\textbf{Model} & \textbf{Classification Error} \\		
		
		Resnext-50 (Euc. \cite{resnext}) & 22.2 \\ 
		Resnext-50 (Euc. $\dagger$) & {\color{red} 22.7} \\ 
		Resnext-50 (Euc.WSS) & {22.3} \\ 
		Resnext-50 (PIO-SOSE)& 21.5 \\ 
		Resnext-50 (PIO-SOSE-WSS)& {\color{blue} 21.3} \\ 
		\midrule 
		
		Mobilenetv2 (Euc. \cite{Sandler})& 28.0 \\ 
		Mobilenetv2 (Euc. $\dagger$)& {\color{red}  27.9} \\ 
		Mobilenetv2 (Euc.-WSS)& {27.5} \\ 
		Mobilenetv2 (PIO-SOSE)& 26.8 \\ 
		Mobilenetv2 (PIO-SOSE-WSS)& {\color{blue} 26.4} \\ 
		\midrule 
		DeepRoots (Euc. \cite{Ioanno})& 26.6 \\ 
		DeepRoots (Euc. $\dagger$)& {\color{red} 27.0} \\ 
		DeepRoots (Euc.-WSS)& {26.6} \\ 
		DeepRoots (PIO-SOSE)& 25.9 \\
		DeepRoots (PIO-SOSE-WSS)& {\color{blue} 25.5} \\ 
		\bottomrule		
		\bottomrule
	\end{tabular}%
	\label{tab:sep}%
\end{table}%

\end{appendices}


\bibliographystyle{unsrt}

\end{document}